\documentclass[a4paper]{cas-dc}
\usepackage{balance}
\usepackage[numbers,sort&compress]{natbib}
\usepackage{amsmath}
\usepackage{wasysym}
\definecolor{customcolor}{RGB}{0, 0, 255} 
\usepackage[numbers]{natbib}
\usepackage{enumitem}
\usepackage{graphicx}
\usepackage{float}
\usepackage{caption}
\usepackage{subcaption}
\usepackage{epstopdf}
\usepackage{threeparttable}
\usepackage{tabularx}
\usepackage{booktabs}
\usepackage{caption}
\usepackage{multicol}
\usepackage{hyperref}
\usepackage{url}
\usepackage{bm}
\usepackage{colortbl}
\hypersetup{
    colorlinks=true,
    linkcolor=customcolor, 
    citecolor=customcolor, 
    filecolor=customcolor, 
    urlcolor=magenta   
}
\usepackage{xcolor,soul}
\usepackage{adjustbox}
\usepackage{algorithmic}
\usepackage{algorithm}
\usepackage[none]{hyphenat}

\definecolor{LightGray}{rgb}{1,1,1}

\def\tsc#1{\csdef{#1}{\textsc{\lowercase{#1}}\xspace}}
\tsc{WGM}
\tsc{QE}

\begin{document}
\let\WriteBookmarks\relax
\def\floatpagepagefraction{1}
\def\textpagefraction{.001}
\let\printorcid\relax

\shorttitle{}    

\shortauthors{L. Lan et~al.}

\title[mode = title]{DMAF-Net: An Effective Modality Rebalancing Framework for Incomplete Multi-Modal Medical Image Segmentation}

\author[1]{Libin Lan}[style=chinese]
\corref{cor1}
\ead{lanlbn@cqut.edu.cn}
\cortext[cor1]{Corresponding author.}

\author[1]{Hongxing Li}[style=chinese]
\ead{hongxing.li@stu.cqut.edu.cn}

\author[1]{Zunhui Xia}[style=chinese]
\ead{zunhui.xia@stu.cqut.edu.cn}

\affiliation[1]{organization={College of Computer Science and Engineering, Chongqing University of Technology},
                city={Chongqing},
              citysep={}, 
                postcode={400054}, 
                state={Chongqing},
                country={China}}

\author[2]{Yudong Zhang}[style=chinese]
\ead{yudongzhang@ieee.org}

\affiliation[2]{organization={School of Computer Science and Engineering, Southeast University},
                city={Nanjing},
              citysep={}, 
                postcode={210096}, 
                state={Jiangsu},
                country={China}}

\begin{abstract}
Incomplete multi-modal medical image segmentation faces critical challenges from modality imbalance, including imbalanced modality missing rates and heterogeneous modality contributions. Due to their reliance on idealized assumptions of complete modality availability, existing methods fail to dynamically balance contributions and neglect the structural relationships between modalities, resulting in suboptimal performance in real-world clinical scenarios. To address these limitations, we propose a novel model, named Dynamic Modality-Aware Fusion Network (DMAF-Net). The DMAF-Net adopts three key ideas. First, it introduces a Dynamic Modality-Aware Fusion (DMAF) module to suppress missing-modality interference by combining transformer attention with adaptive masking and weight modality contributions dynamically through attention maps. Second, it designs a synergistic Relation Distillation and Prototype Distillation framework to enforce global-local feature alignment via covariance consistency and masked graph attention, while ensuring semantic consistency through cross-modal class-specific prototype alignment. Third, it presents a Dynamic Training Monitoring (DTM) strategy to stabilize optimization under imbalanced missing rates by tracking distillation gaps in real-time, and to balance convergence speeds across modalities by adaptively reweighting losses and scaling gradients. Extensive experiments on BraTS2020 and MyoPS2020 demonstrate that DMAF-Net outperforms existing methods for incomplete multi-modal medical image segmentation. Extensive experiments on BraTS2020 and MyoPS2020 demonstrate that DMAF-Net outperforms existing methods for incomplete multi-modal medical image segmentation. Our code is available at \url{https://github.com/violet-42/DMAF-Net}.
\end{abstract}

\maketitle 

\section{Introduction}
\label{sec:introduction}
In clinical practice, multi-modal imaging techniques such as magnetic resonance imaging (MRI) \cite{bakas2017advancing,biondetti2021pet,chen2025magnetic} have been increasingly adopted, where complementary information across modalities can significantly enhance segmentation performance \cite{ding2021rfnet, ma2025dmfusion}. However, acquiring complete MRI datasets in real-world scenarios remains challenging due to data corruption, image degradation, and motion-related artifacts \cite{krupa2015artifacts,graves2013body,havaei2016hemis}. Consequently, despite remarkable progress in multi-modal medical segmentation \cite{wang2023a2fseg,shi2024passion}, its practical deployment continues to face obstacles under missing-modality conditions.

Current research on incomplete multi-modal segmentation predominantly focuses on three methodological directions. The most straightforward approach leverages generative adversarial networks \cite{goodfellow2020generative} or diffusion-based inpainting \cite{meng2024multi,muller2023multimodal} to synthesize missing modalities using the complete cross-modal data, enabling full-modality segmentation through synthetic image generation \cite{zhang2024unified,meng2024multi,zhang2024brain}. Another common strategy employs knowledge distillation to transfer diagnostic patterns from comprehensively trained teacher models (using full modalities) to specialized student models, enabling their adaptation to modality-absent scenarios \cite{wang2023prototype,wang2023learnable}. The current state-of-the-art paradigm adopts shared representation learning, which trains unified segmentation models through coordinated spatial mapping across modalities. This architecture combines modality-specific encoders with a shared decoder \cite{zhang2022mmformer,li2024deformation,wang2023multi}, achieving superior operational flexibility while maintaining computational efficiency compared to the two aforementioned research lines.
\begin{figure*}[!htb] 
		\centering
		\includegraphics[width=1.0\textwidth]{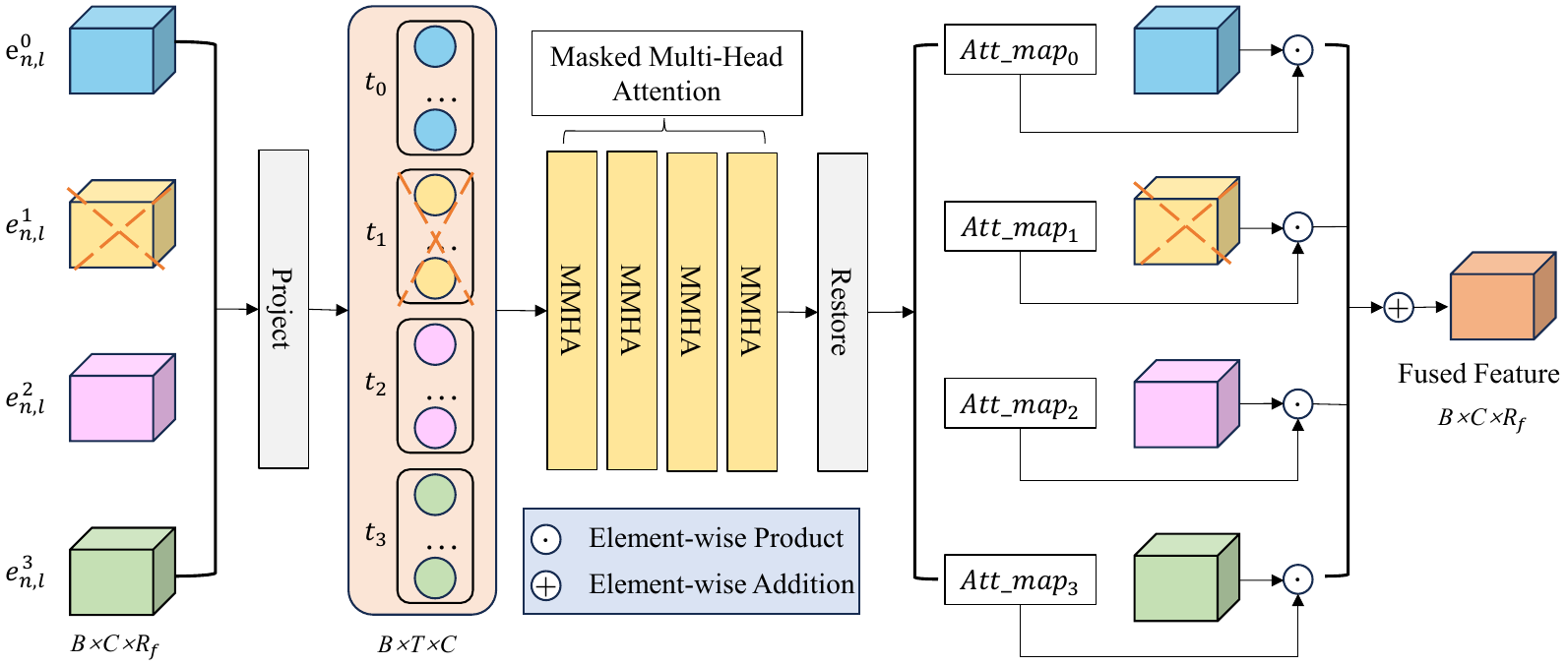}
        \caption{Our proposed Dynamic Modality-Aware Fusion (DMAF) module. Here, $R_f$ denotes the feature dimensions ($D\times H\times W$ for 3D volumes or $H\times W$ for 2D slices), while $T$ represents the tokenized form (with $DHW$ or $HW$ tokens) suitable for attention computation in the transformer architecture. $e_{n,l}^m$ indicates the output feature of the $l$-th layer in encoder $m$.}
		\label{fig:DMAF}
\end{figure*}

Unfortunately, existing works typically prioritize optimizing segmentation performance while oversimplifying training conditions, thereby neglecting the modality imbalance inherent in multi-modal tasks. In clinical practice, this imbalance manifests as two interdependent challenges: imbalanced modality missing rates and heterogeneous modality contributions, which complicate the development of robust and generalizable models.

\textbf{Imbalanced modality missing rates} refers to the unequal absence rates across modalities in real-world scenarios. For instance, in MRI modalities, T1-weighted sequences exhibit lower missing rates due to their robustness against motion artifacts, while T2-weighted sequences suffer higher absence rates attributed to susceptibility to patient movement and elevated acquisition costs \cite{graves2013body}. This disparity causes models to prioritize learning features from high-frequency modalities, resulting in inadequate representation of low-frequency modalities. Dominant modalities not only dominate gradient updates but also suppress the learning processes of other modalities, leading to heterogeneous convergence rates \cite{wei2024fly,peng2022balanced,shi2024passion,fan2023pmr}. Consequently, this undermines the potential superiority of multi-modal learning, causing them to underperform uni-modal approaches in specific diagnostic scenarios — a phenomenon termed modality competition \cite{yang2024facilitating}. Notably, existing modality rebalancing techniques predominantly assume Perfect Data Training (PDT) conditions, where modalities share identical missing rates. Typical PDT implementations either randomly mask modalities per training epoch (ensuring cyclical visibility) or pre-mask modalities before training. However, such idealized assumptions conflict with Imperfect Data Training (IDT) scenarios — the more clinically relevant case where modalities exhibit heterogeneous missing rates and remain persistently invisible once masked.

\textbf{Heterogeneous modality contributions} arises from the unequal diagnostic relevance across modalities. While existing methods implicitly assume equal importance among modalities, their actual contributions to segmentation tasks vary significantly \cite{zhang2024multimodal,wei2024enhancing,zhang2024unleashing,fan2023pmr}. This oversight results in critical pathological features being underdeveloped or distorted within the embedding space, thereby compromising fusion performance. For example, certain modalities contain richer discriminative information for specific structures (e.g., tumor cores), yet current frameworks fail to utilize this potential due to inflexible weighting mechanisms.

Recent works have made notable progress in modality rebalancing \cite{shi2024passion,wei2024fly}. However, they still exhibit several limitations. First, static fusion mechanisms (e.g., feature concatenation \cite{ding2021rfnet}) cannot dynamically suppress missing-modality noise or adaptively weight contributions. Second, traditional distillation methods \cite{wang2023learnable} only partially address imbalance by mimicking the teacher model's output probability distributions (soft labels) while neglecting cross-modal semantic structural relationships. Finally, some approaches employ loss reweighting or gradient scaling techniques to balance convergence rates across modalities, but they primarily focus on classification tasks rather than segmentation \cite{sun2024redcore,fan2023pmr}.

To address these challenges, we propose the \textbf{Dynamic Modality-Aware Fusion Network (DMAF-Net)} for incomplete multi-modal segmentation under imbalanced missing rates. We first propose the Dynamic Modality-Aware Fusion (DMAF) module that masks missing modalities while dynamically weighting modality-specific contributions to enhance the quality of subsequent multi-modal fusion features, as illustrated in Fig. \ref{fig:DMAF}. Building on this foundation, we develop a synergistic framework integrating Relation Distillation and Prototype Distillation for multi-modal segmentation. The Relation Distillation module combines Covariance Consistency Alignment — which preserves high-order statistical relationships through covariance matrix similarity constraints — with Masked Attention Alignment that enables localized cross-modal feature interactions, achieving robust global-local knowledge transfer in modality-missing scenarios. The Prototype Distillation component introduces class-specific semantic prototypes to align uni-modal and fused features in a shared embedding space, preventing model from being biased toward dominant modalities while maintaining semantic consistency across missing modalities via prototype-anchored regularization. Finally, we propose the Dynamic Training Monitoring (DTM) strategy, which dynamically tracks modality-specific distillation gaps and adaptively adjusts loss weights and gradient scaling factors to balance convergence rates across modalities during training. Our contributions are threefold: 
\begin{itemize}
    \item[1.] We propose a Dynamic Modality-Aware Fusion (DMAF) module that integrates transformer-based attention mechanisms with dynamic masking strategies. This design enables adaptive weighting of modality contributions while suppressing interference from missing modalities, overcoming the limitations of static fusion approaches in existing methods.
    \item[2.] We introduce a dual-level distillation framework that synergistically combines Relation Distillation and Prototype Distillation, achieving complementary alignment of global statistics and local details while enforcing semantic consistency across modalities via class-specific prototype alignment.
    \item[3.] We devise a Dynamic Training Monitoring (DTM) strategy that tracks real-time modality-specific distillation gaps. Through adaptive loss reweighting and gradient direction constraints, this system dynamically balances optimization paces across modalities with heterogeneous missing rates, addressing the critical gap in training stability for imbalanced scenarios.
\end{itemize}

This paper is organized as follows. Section \ref{sec2} reviews recent advances in incomplete multi-modal segmentation and modality rebalancing. Section \ref{sec3} presents a detailed description of the proposed DMAF-Net architecture, including problem formulation, dynamic modality-aware fusion, relation and prototype distillation, and dynamic training monitoring strategy. Finally, Section \ref{sec4} presents experimental results, and Section \ref{sec5} concludes this paper, discusses the limitations of the current method, and outlines future work.

\section{RELATED WORK}\label{sec2}
\subsection{Incomplete Multi-modal Segmentation}
Current approaches for incomplete multimodal segmentation fall into three categories: \textbf{generative completion}, \textbf{knowledge distillation}, and \textbf{shared representation learning}. Generative completion methods typically employ GANs or diffusion models \cite{ho2020denoising} to reconstruct missing modalities. For example, M2DN \cite{meng2024multi} proposes a diffusion model-based joint denoising framework to iteratively restore the latent distribution of missing modalities, while SMIL \cite{ma2021smil} introduces Bayesian meta-learning to guide cross-modal feature reconstruction using uncertainty estimation. However, these methods rely on idealized assumptions of complete modality data and often introduce synthetic artifacts \cite{zhang2023motion,ahmad2022new} in high missing rate scenarios, degrading downstream segmentation performance. Knowledge distillation strategies enhance uni-modal inference by transferring supervision signals from full-modality models. ProtoKD \cite{wang2023prototype} and PDKD \cite{wu2024prototype} achieve knowledge transfer by computing class prototypes and leveraging intra-class/inter-class feature similarity relationships. MMANet \cite{wei2023mmanet} designs boundary-aware distillation with modality-aware regularization. Nevertheless, existing approaches remain constrained by the idealized assumption of full-modality teacher models, struggling to address knowledge degradation under dynamic missing patterns. Shared representation learning frameworks, such as RFNet \cite{ding2021rfnet} and mmFormer \cite{zhang2022mmformer}, construct unified feature spaces via multi-encoder and shared-decoder architectures. Nevertheless, their reliance on static fusion mechanisms — such as simple feature concatenation — fails to dynamically balance modality contributions, leading to suboptimal cross-modal semantic coherence. 

A critical limitation of existing studies lies in their exclusive focus on inference-stage modality absence while neglecting the cascading effects of training-phase modality imbalance (e.g., heterogeneous missing rates). This oversight exacerbates representation degradation, particularly for modalities with high missing rates \cite{shi2024passion}.

\subsection{Imbalanced multimodal learning}
Incomplete multimodal segmentation faces dual imbalances: \textbf{imbalanced modality missing rates} and \textbf{heterogeneous modality contributions}. For the former, PASSION \cite{shi2024passion} addresses the challenges of modality imbalance through self-distillation and dynamic balancing mechanisms. RedCore \cite{li2024deformation} combines a variational information bottleneck with relative advantage-aware optimization to suppress feature shifts in high-missing-rate modalities through latent space constraints. ModDrop \cite{xiao2020audiovisual} addresses the issue of inconsistent training speeds among different modalities by randomly dropping paths of distinct modalities, thereby mitigating their overfitting tendencies. For heterogeneous modality contributions, PMR \cite{fan2023pmr} alleviates the issue of dominant modality suppression in multimodal learning through a prototype-guided modality rebalancing mechanism. SFusion \cite{wang2023a2fseg} designs a self-attention dynamic weighting module, HEALNet \cite{yang2024facilitating} enhances cross-modal complementarity with hybrid early fusion, and OGM-GE \cite{peng2022balanced}  further balances optimization speeds via gradient editing. 

Despite these advancements, key challenges remain in multimodal medical image segmentation. conventional distillation methods \cite{shi2024passion,wang2023prototype,dou2020unpaired} overemphasize mimicking output probabilities while neglecting structured relationships critical for cross-modal knowledge transfer, such as spatial dependencies and pathological topological patterns \cite{salguero2024data,wang2025distilling}. To address this, we propose a synergistic framework integrating Relation Distillation and Prototype Distillation. The Relation Distillation component combines covariance-based statistical alignment for global structure preservation with masked graph attention for localized cross-modal interactions, while Prototype Distillation enforces semantic consistency through cross-modal class-specific prototype alignment. This unified global-local paradigm dynamically balances statistical relationships and semantic representations, overcoming the limitations of traditional single-level distillation by enhancing robustness in modality-missing scenarios while mitigating model bias toward dominant modalities.

Beyond these limitations, Current fusion methods for handling missing modalities adopt two primary strategies: 1) \textbf{Partial fusion of available modalities} \cite{wang2023a2fseg,shi2023mftrans}, which avoids invalid data but causes information fragmentation and poor generalization under unseen modality-missing scenarios; 2) \textbf{Zero-padding} \cite{shi2024passion,zhang2022mmformer} to maintain fixed input dimensions, which simplifies the model but introduces noise from padded values. 

To address these limitations, we propose a Dynamic Modality-Aware Fusion (DMAF) module that combines zero-padding with an attention masking mechanism. This method dynamically masks missing modalities during attention computation to suppress noise, retain cross-modal synergy, and address heterogeneous contributions from different modalities while remaining robust to incomplete inputs.

\subsection{Dynamic Training Monitoring Techniques}
Dynamic training monitoring techniques aim to address the core challenges of modality missing rate heterogeneity and optimization imbalance in multimodal medical image segmentation. DynCIM \cite{qian2025dyncim} proposes a dynamic curriculum mechanism that progressively increases the training difficulty for high-missing-rate modalities to alleviate discrepancies in optimization speeds across modalities. Some works, such as \cite{yang2024facilitating,fan2023pmr}, mitigate modality imbalance through loss reweighting, while \cite{wei2024fly,shi2024passion,li2023boosting} balance convergence rates of different modalities by adjusting their training gradients. Diagnosing \& Re-learning \cite{wei2024diagnosing} introduces a dynamic adjustment strategy based on separability diagnosis, dynamically tuning the reinitialization strength of uni-modal encoders. However, these methods largely rely on predefined heuristic rules or local diagnostic signals, failing to globally optimize dynamic knowledge complementarity among modalities. This limitation makes them prone to local optima, especially under high-missing-rate modalities. 

To address these challenges, we propose a Dynamic Training Monitoring (DTM) strategy, which dynamically tracks modality-specific distillation gaps using Exponential Moving Average (EMA) \cite{morales2024exponential}. This enables adaptive loss reweighting and gradient scaling to decelerate the convergence of dominant modalities while accelerating the convergence of underrepresented ones.

\begin{figure*}[!t] 
    \centering
    \includegraphics[width=1.0\textwidth]{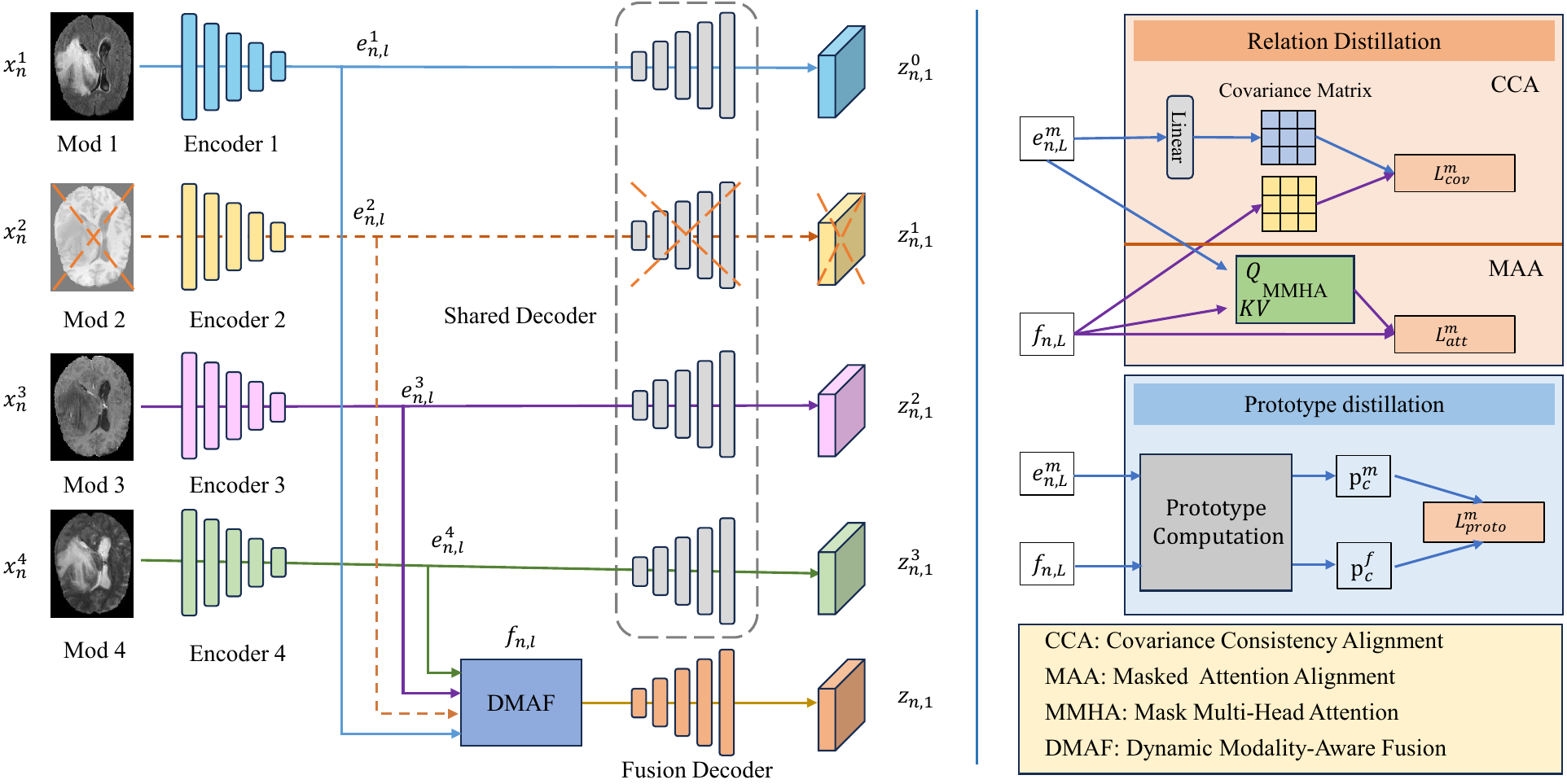}
    \caption{Architecture of our proposed DMAF-Net. The framework processes multi-modal inputs through modality-specific encoders to extract individual feature representations. These features are processed via two distinct decoding pathways: (1) A shared uni-modal decoder that independently handles each modality's features, and (2) a multi-modal fusion pathway where encoder outputs are dynamically integrated at each level through Dynamic Modality-Aware Fusion (DMAF) module before decoding. The system is optimized via complementary loss functions -$\mathcal{L}_{cov}^m$ and $\mathcal{L}_{attn}^m$ for Relation Distillation that enforces global and local structural consistency between uni-modal encoders and fused features, and $\mathcal{L}_{ptoto}^m$ for Prototype Distillation that constrains category-level semantic alignment between uni-modal features $\bm{e}_{n,L}^m$ and fused features $\bm{f}_{n,L}$.}
    \label{fig:framework}
\end{figure*}

\section{METHODS}\label{sec3}
\subsection{Problem Definition}
In incomplete multi-modal medical image segmentation tasks, we consider a multi-modal dataset comprising $N$ samples. Each sample consists of $M$ modalities, where each modality contains $P$ pixels denoted as $\bm{x}_{n,p}^m$. Here, the indices are defined as $n\in[1,2,...N], m\in[1,2,...,M], p\in[1,2,...,P]$. All modalities within the same sample share identical segmentation labels, i.e., $\bm{y}_{n,p}=\bm{y}_{n,p}^{1}=\bm{y}_{n,p}^{2}=...=\bm{y}_{n,p}^{M}$. To characterize modality absence, we introduce a binary modality presence matrix $\bm{I}\in\{0,1\}^{N\times M}$, where $\bm{I}_{n,m}=1$ indicates the existence of modality m in sample n, otherwise it equals $0$. The missing rate for modality $m$ is formulated as $\bm{MR}^m=(N- {\textstyle \sum_{n\in [N]}} \bm{I}_{n,m})/N$, where $\bm{MR}^m\in[0,1)$ due to the constraint that each sample contains at least one modality.

State-of-the-art approaches \cite{shi2024passion,zhang2022mmformer} typically adopt a multi-encoder dual-decoder architecture. This architecture includes modality-specific encoders $\{E_m\}_{m=1}^{M}$ paired with complementary decoding pathways: a shared decoder $D_{s}$ for refining modality-exclusive features and a fusion-aware decoder $D_f$ designed for cross-modal integration. The framework implements hierarchical feature propagation through U-Net-style skip connections, allowing multi-scale representations from accessible encoders to be progressively fused via both decoders. In this architecture, the fused multi-modal features at layer $l$ are formalized as $\bm{z}_{n,l}=D_{f,l}(\bm{x}_n)$, while the modality-specific representations are defined as $\bm{z}_{n,l}^m = D_{s,l}(\bm{x}_n^m)$, with $l=1$ denoting the final output layer. The overarching objective of this paradigm is
\begin{equation}
\label{eq:01}
\scalebox{0.92}{$
\mathcal{L}_{\mathrm{task}} = \underbrace{\sum_{l=1}^{L}\lambda_l\ell_{\mathrm{dice+ce}}(\mathrm{Up}_{2^{l}}(\bm{z}_{n,l}),\bm{y}_n)}_{\mathcal{L}_{\mathrm{fuse}}} + \underbrace{\sum_{m=1}^{M}I_{n,m}\ell_{\mathrm{dice+ce}}(\bm{z}_{n,1}^m,\bm{y}_n)}_{\mathcal{L}_{\mathrm{sep}}},
$}
\end{equation}
where $\ell_{dice+ce}$ denotes the combination of Dice loss and weighted cross-entropy loss, $\lambda_l=1/2^l$ is the hierarchical loss weight, and $\mathrm{Up}_{2^{l}}$ represents upsampling by a factor of $2^{l}$. Through imposing deep supervision on the fused encoder, $\mathcal{L}_{\mathrm{fuse}}$ enhances the model's feature representation capability and training efficiency; Simultaneously, $\mathcal{L}_{\mathrm{sep}}$ enforces the model to retain unique information from each modality by independently supervising each modality's segmentation results, thereby enhancing uni-modal robustness and accuracy of multi-modal fusion.

However, facing the issue of modality imbalance, existing methods encounter two key challenges. First, during the process of multi-modal feature fusion, simply concatenating or adding features fails to dynamically weigh the contribution of each modality, leading to dominant modalities excessively dominating the fusion process. Second, $\mathcal{L}_{\mathrm{sep}}$ is primarily controlled by dominant modalities, causing high-missing-rate modalities to suffer from feature degradation due to insufficient gradient updates, exacerbating the disparity in optimization speeds among modalities, which may lead to severe fusion issues within state-of-the-art paradigms. These limitations significantly restrict the generalization ability of models in real clinical scenarios, motivating us to balance the performance of each modality in a unified end-to-end training framework, considering the imbalanced modality missing rates. The architecture of our proposed DMAF-Net is depicted in Fig. \ref{fig:framework}.

\subsection{Dynamic Modality-Aware Fusion}
In multi-modal medical image segmentation tasks, the amount of discriminative information and task relevance inherent in different modalities vary, directly affecting each modality's contribution to the final segmentation performance. To dynamically adapt to scenarios with missing modalities and fully utilize the information from available modalities, we propose a Dynamic Modality-Aware Fusion (DMAF) module. By integrating transformer attention mechanisms with dynamic masking strategies, this module achieves adaptive fusion of cross-modal features while effectively suppressing the interference caused by missing modalities on the model, as illustrated in Fig. \ref{fig:DMAF}. The DMAF includes the following key steps:

\textbf{Feature Processing}: For the multi-modal features $\bm{e}_{n,l}^{m}$ output from the $l$-th layer of the encoder, we first apply a dynamic mask $\bm{C}_{n,m}$ to zero out features of missing modalities, retaining only those from effective modalities to ensure that invalid modalities do not interfere with subsequent computations. Subsequently, adaptive convolutional downsampling is used to unify the resolution of each modality's features to a fixed size $\bm{S}_d=H_d\times W_d\times D_d$, significantly reducing the computational complexity for subsequent attention mechanisms.

\textbf{Cross-Modal Attention Learning}: The downsampled features are flattened into token sequences $\{\bm{t}_m\in\mathbb{R}^{S_d\times C}\}$, concatenated, and fed into $K$ layers of MMHA modules. The computation at the $k$-th layer of MMHA is given by concatenating tokens from different modalities into a tensor $\bm{T}_0\in\mathbb{R}^{(S_d\cdot M)\times C}$ suitable for transformer input format, then processing through the MMHA module to learn latent multi-modal correlations. Specifically, the $k$-th layer MMHA is expressed as:
\begin{equation}
    \begin{split}
		\bm{T}_{k}^{\prime}&=\mathrm{MMHA}(\mathrm{LN}(\bm{T}_{k-1}),\bm{M}_{mask})+\bm{T}_{k-1},\\
        \bm{T}_{k}&=\mathrm{FFN}(\mathrm{LN}(\bm{T}_{k}^{\prime}))+\bm{T}_{k}^{\prime},
    \end{split}
    \label{eq:02}
\end{equation}
where $\bm{M}_{mask}\in\{-\infty,0\}^{S_d\cdot M}$ is an attention mask dynamically generated according to $\bm{C}$, compelling the model to disregard tokens from missing modalities, thereby preventing the introduction of noise interference.

\textbf{Feature Re-weighting Fusion}: The tokens $\bm{T}_{K}^{m}$ output from the DMSA for each modality undergo de-serialization and upsampling, followed by a Softmax operation to generate spatial weight maps $\bm{a}_m\in[0,1]^{H_l\times W_l\times D_l}$. The final fused feature is calculated as:
\begin{equation}
    \bm{f}_{n,l}=\sum_{m=1}^{M} \bm{e}_{n,l}^{m}\odot {\bm{a}_m},
    \label{eq:03}
\end{equation}
where $\odot$ denotes element-wise multiplication. This mechanism enables the model to dynamically focus on discriminative regions while suppressing noise interference.

\subsection{Relationship and Prototype Distillation}
In incomplete multi-modal medical image segmentation, the over-dominance of low-missing-rate modalities significantly impedes effective learning from high-missing-rate modalities. Existing knowledge distillation methods \cite{shi2024passion,wang2023prototype,dou2020unpaired} transfer multi-modal fusion information to uni-modal models. Specifically, for dominant modalities, their features are closer to the multi-modal fused features, resulting in smaller distillation losses. In contrast, for other modalities with lower similarity to the fused features, the distillation losses are higher, forcing these low-frequency modalities to learn from the fused modality and thereby achieving modality rebalancing. However, these methods perform knowledge distillation only at the final output layer of the decoder (e.g., segmentation predictions), focusing solely on the similarity between uni-modal and multi-modal models in terms of final prediction results. They fail to capture feature-level relationships, lack statistical alignment and semantic consistency, and hinder uni-modal models from learning deep multi-modal knowledge, especially under missing-modality conditions. To address this issue, we propose a synergistic relation-prototype distillation framework that constrains the interaction between uni-modal and fused features through two key mechanisms: feature statistical Relation Distillation and semantic Prototype Distillation.

\subsubsection{Relation Distillation}
Relation Distillation, through two key components — Covariance Consistency Alignment (CCA) and Masked Attention Alignment (MAA) — operating at the final layer of the encoder, enforces consistency in the relationship between fused features and uni-modal features.

\textbf{CCA} directly computes spatial covariance matrices of fused features $f_{n,l}$ and uni-modal features $e_{n,L}^{m}$:
\begin{equation}
    \mathrm{Cov}(\mathbf{X})=\frac{1}{HWD}(\mathbf{X}-\mu)^T(\mathbf{X}-\mu),
    \label{eq:04}
\end{equation}
where $\mathbf{X}\in\mathbb{R}^{C\times H\times W\times D}$ denotes the fused multimodal features or the unimodal features, $\mu$ represents the mean of the feature. The covariance matrix $\mathrm{Conv(\cdot)}$ captures the statistical correlations between features, while the uni-modal covariance is projected into the fused feature space via a linear mapping $\mathcal{P}$ for alignment. This process enforces the uni-modal model to learn the distribution patterns of multi-modal features, constrained by the mean squared error:
\begin{equation}
    \mathcal{L}_{\mathrm{cov}}^m=\left\|\mathcal{P}(\mathrm{Cov}(\bm{e}_{\mathrm{n,L}}^m)) - \mathrm{Cov}(\bm{f}_{\mathrm{n,L}})\right\|_{2}^2.
    \label{eq:05}
\end{equation}

This covariance alignment mechanism effectively guides uni-modal branches to learn the global statistical characteristics of multi-modal features. Through explicit constraints, it reduces the dependence of uni-modal models on other modalities (e.g., retaining multi-modal statistical properties even when certain modalities are missing), mitigates feature distribution shifts caused by missing modalities, and enhances the model's generalization ability under incomplete modality conditions.

\textbf{MAA} employs the uni-modal features $e_{n,l}^{m}$ as the query (Q), while using fused multi-modal feature $f_{n,L}$ as key-value pairs (K,V). Through the Multi-Head Attention (MHA) mechanism, this architecture generates attention weight matrices that dynamically capture cross-modal interactions:
\begin{equation}
    \bm{A}^m=\mathrm{MHA}(Q=\bm{e}_{n,L}^m,K=\bm{f}_{n,L},V=\bm{f}_{n,L},\bm{M}_{mask}),
    \label{eq:06}
\end{equation}
where $\bm{M}_{mask}\in\{-\infty,0\}$ is an attention mask generated based on the modality presence matrix $C$, forcing the model to ignore key-value pairs corresponding to missing modalities. By minimizing the MSE loss between attention outputs and fused features, the model captures the attention patterns of the multi-modal model in critical regions (e.g., lesion boundaries), ensuring that the uni-modal model focuses on the same important areas as the multi-modal model:
\begin{equation}
    \mathcal{L}_{\mathrm{attn}}^m=\left\|\bm{A}^m -\bm{f}_{n,L}\right\|_2^2.
    \label{eq:07}
\end{equation}

The total Relation Distillation loss is computed as the average of covariance and attention losses across all valid modalities:
\begin{equation}
    \mathcal{L}_{\mathrm{rel}}=\frac{1}{|M|}\sum_{m=1}^{M} (\alpha_1\mathcal{L}_{\mathrm{conv}}^m+(1-\alpha_1)\mathcal{L}_{\mathrm{attn}}^m),
    \label{eq:08}
\end{equation}
where $\alpha_1$ is a learnable balancing hyperparameter that undergoes dynamic updates during training.

In Relation Distillation, the covariance mechanism ensures consistency between uni-modal features and multi-modal fused features in terms of global statistical properties, enhancing the model's robustness to distribution shifts. The attention mechanism captures the attention dependencies of the multi-modal model in critical regions, strengthening local feature alignment. Through complementary constraints at both global and local levels, comprehensive knowledge transfer from multi-modal to uni-modal models is achieved, enabling the uni-modal model to inherit the global structure and local details of the multi-modal model even when other modalities are missing.

\subsubsection{Prototype Distillation}
The prototype Distillation module focuses on enhancing cross-modal consistency of categorical semantics. For category $c$, the fused prototype $\bm{p}_c^f$ and uni-modal prototype $\bm{p}_c^m$ are computed as:
\begin{equation}
    \bm{p}_{c}^{f}=\frac{\sum_{\bm{x}\in\Omega_{c}}\bm{f}_{n,L}(\bm{x})}{\mid\Omega_{c}\mid+\epsilon},\bm{p}_{c}^{m}=\frac{\bm{I}_{n}^m\cdot\sum_{\bm{x}\in\Omega_{c}}\bm{e}_{n,L}^m(\bm{x})}{\mid\Omega_{c}\mid+\epsilon},
    \label{eq:09}
\end{equation}
where $\Omega_c$ denotes the pixel set of category $c$, and $\epsilon$ prevents division-by-zero errors. A prototype alignment loss is employed to enforce prototype similarity between fused and uni-modal representations:
\begin{equation}
    \mathcal{L}_{\mathrm{proto}}=\frac{1}{|M|}\sum_{m=1}^{M}\sum_{c=1}^{C} \left(1-\mathrm{cos}(\bm{p}_{c}^{f},\bm{p}_{c}^{m})/\tau_m\right),
    \label{eq:10}
\end{equation}
$\tau_m$ is a temperature hyperparameter used to control the optimization speed of different modalities. By explicitly aligning cross-modal categorical prototypes in real-time, this approach compels uni-modal encoders to maintain semantic consistency with fused representations even under modality absence, thereby simultaneously alleviating distribution shift caused by missing data and enhancing generalization capabilities in incomplete multi-modal scenarios.

\subsection{Dynamic Training Monitoring Strategy}
When addressing incomplete multi-modal medical image segmentation tasks with imbalanced missing rates, the substantial variation in modality absence ratios across different imaging types critically compromises training stability and model performance. To mitigate this challenge, we propose a Dynamic Training Monitoring (DTM) Strategy that achieves modality rebalancing through real-time monitoring of two key indicators: 1) Relation Distillation gaps quantifying feature interaction discrepancies, and 2) Prototype Distillation gaps measuring semantic alignment deviations. By dynamically adjusting both loss weights and gradient scaling factors across modalities, DTM effectively decelerates the convergence speed of dominant modalities while accelerating under-trained ones, thereby establishing balanced parameter optimization dynamics. 

The computational formulations for the Relation Distillation Gap and Prototype Distillation Gap are systematically defined as follows:
\begin{equation}
    \begin{split}
		\bm{g}_{r}(m)&=\alpha_1 \mathcal{L}_{\mathrm{cov}}^m+(1-\alpha_1)\mathcal{L}_{\mathrm{attn}}^m,\\
        \bm{g}_{p}(m)&=\left(1-\mathrm{cos}(\bm{p}_{c}^{f},\bm{p}_{c}^{m}\right)/\tau_m.
    \end{split}
    \label{eq:11}
\end{equation}

To maintain training stability, we employ Exponential Moving Average (EMA) \cite{morales2024exponential} to compute the Relational Distillation gap $\bm{g}_{r}^t(m)$ and Prototype Distillation gap $\bm{g}_{p}^t(m)$ for modality $m$ at timestep $t$, with the smoothed gap estimates updated via EMA formulation:
\begin{equation}
    \begin{split}
	\bm{g}_{r}^{t}(m)&=\alpha_{decay} \bm{g}_{r}^{t-1}(m)+(1-\alpha_{decay}) \bm{g}_{r}^{t}(m), \\
\bm{g}_{p}^{t}(m)&=\alpha_{decay} \bm{g}_{p}^{t-1}(m)+(1-\alpha_{decay}) \bm{g}_{p}^{t}(m),
    \end{split}
    \label{eq:12}
\end{equation}
where $\alpha_{decay}$ is an adaptive decay coefficient computed as:
\begin{equation}
    \begin{split}
	\mathrm{ratio }_{m} & =\frac{\bm{g}_{r}^{(t-1)}(m)+\epsilon }{\bm{g}_{p}^{(t-1)}(m)+\epsilon }, \\
\alpha_{\text {decay }} & =0.9 \cdot\left(1-\operatorname{S}(\mathrm{ratio }_{m})\right),
    \end{split}
    \label{eq:13}
\end{equation}
where $\operatorname{S}(\cdot)$ represents sigmoid function, $\epsilon$ prevents division-by\-zero errors. The decay rate adapts to the historical gap ratio between relation and prototype losses, ensuring faster tracking when gaps diverge significantly.

The total gap $\bm{g}_{\text{total}}^t(m)$ is computed via weighted summation:
\begin{equation}
    \bm{g}_{total}^{t}(m)=\alpha_2 \bm{g}_{r}^{t}(m)+(1-\alpha_2) \bm{g}_{p}^{t}(m),
    \label{eq:14}
\end{equation}
where $\alpha_2$ is dynamically calibrated as the EMA of historical relation-to-total gap ratios:
\begin{equation}
    \alpha_{2}=\frac{\overline{\bm{g}_{r}}}{\overline{\bm{g}_{r}}+\overline{\bm{g}_{p}}} , \quad \bar{\bm{g}.}=\text { EMA of } \bm{g}.
    \label{eq:15}
\end{equation}

To enhance robustness against modality characteristic variations, we introduce a counteractive weighting mechanism that inversely scales modality contributions based on their temporal stability patterns:
\begin{equation}
    \bm{w}^{(t)}(m)=\frac{1 / \bm{g}_{\text {total }}^{(t)}(m)}{\sum_{m^{\prime} \in \mathcal{M}} 1 / \bm{g}_{\text {total }}^{(t)}\left(m^{\prime}\right)}.
    \label{eq:16}
\end{equation}

These weights are applied to the uni-modal segmentation loss $\mathcal{L}_{sep}^m$, with the weighting formulation defined as:
\begin{equation}
    \mathcal{\hat L}_{sep}^m=\bm{w}^{t}(m)\mathcal{L}_{sep}^m.
    \label{eq:17}
\end{equation}

For encoder $m$, the gradient update mechanism operates as follows:
\begin{equation}
    \begin{aligned}
&\gamma^{(t)}(m) =\operatorname{clip}\left(\frac{1}{\bm{w}^{(t)}(m)}, 0.1,10.0\right), \\
&\operatorname{sim}^{(t)}(m) =\cos \left(\nabla_{\theta_{m}}^{(t)}, \nabla_{\theta_{m}}^{(t-1)}\right), \\
&\nabla_{\theta_{m}}^{(t)} =\left\{\begin{array}{ll}
0.7 \gamma^{(t)}(m) \cdot \nabla_{\theta_{m}}^{(t)}, & \text { if } \operatorname{sim}^{(t)}(m)<-0.5 \\
\gamma^{(t)}(m) \cdot \nabla_{\theta_{m}}^{(t)}, & \text { otherwise },
\end{array}\right.
\end{aligned}
    \label{eq:18}
\end{equation}
where $\theta_{m}^{(t)}$ denotes the encoder gradient of modality $m$ at timestep $t$, $\gamma^{(t)}(m)$ represents the gradient scaling factor inversely proportional to the modality's distillation gap, and $\operatorname{sim}^{(t)}(m)$ measures the cosine similarity between consecutive gradient directions. When $\operatorname{sim}^{(t)}(m)$ falls below -0.5, indicating conflicting gradient directions across consecutive iterations, the update magnitude is adaptively reduced by a factor of 0.7 to suppress optimization instability. This specific coefficient was determined through comprehensive grid search experimentation over candidate values \{0.5, 0.6, 0.7, 0.8\}, with 0.7 demonstrating optimal stabilization while maximally preserving critical gradient information essential.

\subsection{Overall Objective}
The overall loss function of our proposed DMAF-Net is:
\begin{equation}
    \mathcal{L}=\lambda_1 \mathcal{L}_{fuse}+\lambda_2 \mathcal{L}_{sep}+\lambda_3 \mathcal{L}_{rel}+\lambda_4 \mathcal{L}_{proto},
    \label{eq:19}
\end{equation}
where $\lambda_1, \lambda_2, \lambda_3, \lambda_4$ are balancing hyperparameters.

\begin{table*}
    \caption{Quantitative comparison on BraTS2020 and MyoPS2020 under various settings. $MR$ is short for the missing rates of modalities (T1/T1c, Flair, T2) for BraTS2020 and (bSSFP, LGE, T2) for MyoPS2020 respectively. $s$, $m$, and $l$ correspond to small, medium, and large, set as $s=0.2$, $m=0.5$, and $l=0.8$ for BraTS2020 and $s=0.3$, $m=0.5$, and $l=0.7$ for MyoPS2020, respectively. $PDT$ denotes balanced modality distributions $MR=(0,0,0)$ for perfect data training. Here, T1 and T1c share the same missing rate but are treated as separate modalities.}
    \label{tab:brats&myo}
    \resizebox{\textwidth}{!}{
    \begin{tabular}{c|l|cccc|cccc|cccc|cccc}
        \hline
        \multirow{3}{*}{$MR$}      & Dataset  & \multicolumn{8}{c|}{BraTS2020}                                                                                                         & \multicolumn{8}{c}{MyoPS2020}         \\
        \cline{2-18} 
        & Metric   & \multicolumn{4}{c|}{DSC {[}\%{] $\uparrow$}}     & \multicolumn{4}{c|}{HD {[}mm{]} $\downarrow$}    & \multicolumn{4}{c|}{DSC {[}\%{]} $\uparrow$}                                  & \multicolumn{4}{c}{HD {[}mm{]} $\downarrow$}                                   \\
        \cline{2-18} 
        & Method   & WT   & TC     & ET     & Avg.   & WT     & TC      & ET      & Avg.      & LVB       & RVB       & MYO       & Avg.      & LVB       & RVB       & MYO       & Avg.      \\
        \hline
	$PDT$    
        & Baseline                           & 84.42  & 72.33  & 54.86  & 70.54  & 15.13   & 17.30   & 13.22   & 15.22   & 82.46   & 62.86   & 79.63    & 74.98  & 8.08  & 16.62   & 8.93   & 11.21    \\
	\hline
	\multirow{4}{*}{($s$,$m$,$l$)} 
        & Baseline\cite{zhang2022mmformer}   & 81.85  & 69.42  & 51.89  & 67.63  & 16.78   & 16.77   & 12.53   & 15.36   & 77.43   & 56.74   & 71.22    & 68.46  & 19.78  & 23.89  & 21.72  & 21.80    \\
	& +ModDrop\cite{xiao2020audiovisual} & 81.41  & 68.54  & 51.49  & 67.15  & 19.65   & 17.82   & 12.92   & 16.80   & 80.71   & 52.56   & 74.86    & 69.37  & 15.93  & 31.33  & 20.43  & 22.56    \\
	& +PMR\cite{fan2023pmr}              & 82.26  & 70.06  & 52.68  & 68.33  & 17.02   & 18.51   & 13.35   & 16.29   & 78.27   & 58.36   & 73.42    & 70.02  & 18.47  & 25.13  & 17.54  & 20.38    \\
	& +PASSION\cite{shi2024passion}      & 83.41  & 70.46  & 52.41  & 68.76  & 12.85   & 12.94   & 11.10   & 12.29   & 80.03   & 61.21   & 74.57    & 71.94  & 13.22  & 24.46  & 14.47  & 17.38    \\
        & +\textbf{DMAF-Net}                     & \textbf{84.35}  & \textbf{71.84}  & \textbf{53.29}  & \textbf{69.83}  & \textbf{11.73}   & \textbf{12.57}  & \textbf{10.14}   & \textbf{11.48}   & \textbf{81.38}   & \textbf{64.07}   & \textbf{76.74}    & \textbf{74.05}  & \textbf{11.01}  & \textbf{22.88}  & \textbf{12.72}  & \textbf{15.54}    \\
	\hline
	\multirow{4}{*}{($s$,$l$,$m$)} 
        & Baseline\cite{zhang2022mmformer}   & 81.66  & 67.61  & 50.10  & 66.46  & 19.43   & 19.20   & 13.93   & 17.52   & 74.70   & 54.83   & 66.08    & 65.20  & 25.29  & 31.44   & 25.24  & 27.32    \\
	& +ModDrop\cite{xiao2020audiovisual} & 80.46  & 69.41  & 51.33  & 67.07  & 25.37   & 24.51   & 18.82   & 22.9    & 78.86   & 56.42   & 73.94    & 69.74  & 16.45  & 25.86   & 17.66  & 19.99    \\
	& +PMR\cite{fan2023pmr}              & 82.34  & 68.62  & 50.65  & 67.20  & 16.24   & 17.56   & 12.47   & 15.42   & 77.65   & 58.01   & 72.95    & 69.54  & 16.52  & 25.30   & 18.83  & 20.22    \\ 
	& +PASSION\cite{shi2024passion}      & 83.49  & 70.83  & 52.82  & 69.05  & 11.52   & 12.66   & 10.92   & 11.7    & 79.24   & 59.35   & 74.71    & 71.15  & 18.03  & 22.57   & 16.37  & 18.99    \\
        & +\textbf{DMAF-Net}                     & \textbf{84.46}  & \textbf{71.73}  & \textbf{53.75}  & \textbf{69.98}  & \textbf{10.59}   & \textbf{12.08}   & \textbf{9.83}    & \textbf{10.83}   & \textbf{80.56}   & \textbf{62.56}   & \textbf{76.85}    & \textbf{73.32}  & \textbf{15.97}  & \textbf{20.67}   & \textbf{14.06}  & \textbf{16.90}    \\
	\hline
	\multirow{4}{*}{($m$,$s$,$l$)} 
        & Baseline\cite{zhang2022mmformer}   & 81.45  & 70.03  & 52.00  & 67.83  & 20.24   & 23.02   & 16.33   & 19.86   & 76.97   & 54.48   & 71.53    & 67.66  & 15.63  & 22.71   & 24.57  & 20.97    \\
	& +ModDrop\cite{xiao2020audiovisual} & 80.53  & 68.60  & 50.29  & 66.47  & 27.35   & 29.62   & 22.82   & 26.60   & 79.68   & 56.66   & 74.70    & 70.35  & 12.86  & 22.63   & 17.33  & 17.61    \\
	& +PMR\cite{fan2023pmr}              & 81.57  & 69.86  & 51.54  & 67.66  & 19.15   & 20.64   & 15.47   & 18.42   & 77.52   & 58.27   & 70.69    & 68.83  & 19.52  & 25.27   & 20.46  & 21.75    \\
	& +PASSION\cite{shi2024passion}      & 82.96  & 69.78  & 52.62  & 68.45  & 11.36   & 13.58   & 12.36   & 12.43   & 79.43   & 65.25   & 71.64    & 72.11  & 12.50  & 22.09   & 17.08  & 17.22    \\
        & +\textbf{DMAF-Net}                     & \textbf{84.02}  & \textbf{70.76}  & \textbf{53.41}  & \textbf{69.40}  & \textbf{10.23}   & \textbf{13.10}   & \textbf{11.25}   & \textbf{11.53}   & \textbf{80.49}   & \textbf{67.82}   & \textbf{73.52}    & \textbf{73.94}  & \textbf{10.46}  & \textbf{20.72}   & \textbf{14.96}  & \textbf{15.38}    \\
	\hline
	\multirow{4}{*}{($m$,$l$,$s$)} 
        & Baseline\cite{zhang2022mmformer}   & 81.42  & 68.50  & 51.96  & 67.29  & 17.50   & 20.63   & 14.90   & 17.68   & 75.47   & 47.92   & 70.34    & 64.56  & 22.80  & 31.07   & 19.09  & 24.32    \\
	& +ModDrop\cite{xiao2020audiovisual} & 80.94  & 68.77  & 50.83  & 66.85  & 24.83   & 25.96   & 20.06   & 23.62   & 76.60   & 51.61   & 73.66    & 67.29  & 22.03  & 25.77   & 17.30  & 21.70    \\
	& +PMR\cite{fan2023pmr}              & 81.87  & 69.28  & 52.49  & 67.88  & 16.74   & 20.67   & 14.51   & 17.31   & \textbf{78.73}   & 51.70   & 70.83    & 67.09  & 20.62  & 24.76   & 23.55  & 22.98    \\
	& +PASSION\cite{shi2024passion}      & 83.46  & 69.64  & 53.03  & 68.71  & 13.05   & 15.35   & 13.20   & 13.87   & 78.68   & 56.08   & 73.13    & 69.30  & 17.46  & 25.81   & 18.37  & 20.55    \\
        & +\textbf{DMAF-Net}                     & \textbf{84.36}  & \textbf{70.73}  & \textbf{53.82}  & \textbf{69.64}  & \textbf{11.96}   & \textbf{15.08}   & \textbf{12.73}   & \textbf{13.26}   & 78.46   & \textbf{59.73}   & \textbf{74.08}    & \textbf{70.76}  & \textbf{15.21}  & \textbf{23.97}   & \textbf{16.29}  & \textbf{18.49}    \\
	\hline
	\multirow{4}{*}{($l$,$s$,$m$)} 
        & Baseline\cite{zhang2022mmformer}   & 80.70  & 68.57  & 50.72  & 66.66  & 17.82   & 17.71   & 13.23   & 16.25   & 77.52   & 52.88   & 70.45    & 66.95  & 18.56  & 29.17   & 19.70  & 22.48    \\
	& +ModDrop\cite{xiao2020audiovisual} & 80.58  & 68.30  & 50.63  & 66.50  & 25.30   & 25.41   & 19.87   & 23.53   & 81.00   & 54.06   & 75.06    & 70.04  & 16.20  & 22.06   & 17.68  & 18.65    \\
	& +PMR\cite{fan2023pmr}              & 80.76  & 68.16  & 51.48  & 66.80  & 15.34   & 16.32   & 12.45   & 14.70   & 79.75   & 55.72   & 71.96    & 69.14  & 16.28  & 26.28   & 23.55  & 22.04    \\
	& +PASSION\cite{shi2024passion}      & \textbf{83.12}  & 69.25  & 52.00  & 68.12  & 12.16   & 14.58   & 12.40   & 13.05   & 80.56   & 57.96   & 74.71    & 71.08  & 14.96  & 25.82   & 17.80  & 19.53    \\
        & +\textbf{DMAF-Net}                     & 82.43  & \textbf{69.93}  & \textbf{52.89}  & \textbf{68.42}  & \textbf{11.06}   & \textbf{14.09}   & \textbf{11.34}   & \textbf{12.16}   & \textbf{81.83}   & \textbf{60.17 }  & \textbf{76.92}    & \textbf{72.97}  & \textbf{13.24 } & \textbf{24.53}   & \textbf{15.62}  & \textbf{17.80 }   \\
	\hline
	\multirow{4}{*}{($l$,$m$,$s$)}
        & Baseline\cite{zhang2022mmformer}   & 80.53  & 68.63  & 52.15  & 67.10  & 17.52   & 18.88   & 15.75   & 17.38   & 80.06   & 51.88   & 74.67    & 68.87   & 13.41  & 24.17  & 15.37  & 17.76    \\
	& +ModDrop\cite{xiao2020audiovisual} & 80.10  & 68.37  & 50.70  & 66.39  & 23.66   & 24.47   & 19.08   & 22.40   & 80.58   & 52.43   & 75.33    & 69.45   & 12.70  & 23.45  & 14.01  & 16.72    \\
	& +PMR\cite{fan2023pmr}              & 81.67  & 67.78  & 50.69  & 66.71  & 18.14   & 19.70   & 14.75   & 17.53   & 80.14   & 55.49   & 75.24    & 70.29   & 12.44  & 24.26  & 14.59  & 17.10    \\
	& +PASSION\cite{shi2024passion}      & 82.55  & 68.41  & 51.79  & 67.58  & 12.70   & 15.68   & 12.46   & 13.61   & 79.91   & 58.35   & 74.03    & 70.76   & 12.26  & 23.71  & 13.95  & 16.64    \\
        & +\textbf{DMAF-Net}                     & \textbf{83.62}  & \textbf{69.86}  & \textbf{52.58}  & \textbf{68.69}  & \textbf{11.64}   & \textbf{15.03}   & \textbf{11.39}   & \textbf{12.69}   & \textbf{81.16}   & \textbf{61.66}   & \textbf{75.60}    & \textbf{72.81}   & \textbf{11.62}  & \textbf{23.02 } & \textbf{13.60}  & \textbf{16.08}    \\
	\hline
    \end{tabular}
		}
\end{table*}

\section{EXPERIMENTS}\label{sec4}
\subsection{Datasets and Evaluation Metrics}
We conduct segmentation experiments on two publicly available MRI datasets, including:
\begin{enumerate}
		
\item \textbf{BraTS2020} \cite{menze2014multimodal}: The BraTS2020 dataset comprises 369 multi-parametric MRI scans, including T1-weighted, contrast-enhanced T1 (T1ce), T2-weighted, and Fluid Attenuated Inversion Recovery (FLAIR) sequences. Each case is meticulously annotated to delineate healthy brain tissue and three tumor sub-regions: necrotic core and non-enhancing tumor (NCR/NET), peritumoral edema (ED), and enhancing tumor (ET). Following the challenge protocol, these sub-regions are aggregated into three clinically relevant targets: whole tumor (WT: NCR+NET+ED+ET), tumor core (TC: NCR+NET+ET), and active tumor (ET). The dataset is partitioned into 219 training cases, 50 validation cases, and 100 test cases.
		
\item \textbf{MyoPS2020} \cite{qiu2023myops,zhuang2018multivariate}: The MyoPS2020 dataset consists of 45 multi-sequence cardiac MRI scans, including balanced Steady-State Free Precession (bSSFP), Late Gadolinium Enhancement (LGE), and T2-weighted sequences. Annotations focus on left ventricular myocardial pathology, originally categorizing six structures: normal myocardium, left/right ventricular blood pools (LVB/RVB), myocardial edema, scar tissue, and background. To streamline clinical analysis, pathological regions (edema and scar) are merged with normal myocardium into a composite label for left ventricular myocardium (MYO). The dataset is divided into 25 training cases with multi-slice annotations and 20 independent test cases for testing. 
		
\end{enumerate}

Following \cite{shi2024passion,ding2021rfnet}, we cropped the black background regions outside the brain, performed center cropping on the cardiac region, and further normalized the intensity of each volume to zero mean and unit variance. To demonstrate the flexibility of DMAF-Net, we conducted 3D segmentation on the BraTS2020 dataset and 2D segmentation on the MyoPS2020 dataset.
	
The evaluation framework employs two principal performance indicators: the Dice Similarity Coefficient (DSC) for volumetric overlap analysis and Hausdorff Distance (HD) for boundary alignment assessment. DSC quantifies pixel-wise segmentation accuracy through set intersection measurement, whereas HD characterizes maximum surface deviation between segmented and ground truth contours. Optimal segmentation quality is reflected by maximal DSC values (ideal score = 1) and minimized HD measurements (ideal value = 0). In accordance with established biomedical segmentation benchmarks like BraTS and MyoPS, all quantitative comparisons were conducted at the whole-volume analysis level to maintain methodological consistency with clinical evaluation protocols.
	
\subsection{Implementation Details}
To establish a comprehensive comparison with state-of\-the-art (SOTA) imbalanced multi-modal learning approaches (ModDrop \cite{xiao2020audiovisual}, PMR \cite{fan2023pmr}, PASSION \cite{shi2024passion}), we employ mmFormer \cite{zhang2022mmformer} as our baseline architecture to evaluate method performance under varying modality missing rate scenarios. Furthermore, we incorporate three additional SOTA multi-modal segmentation frameworks — RFNet \cite{ding2021rfnet}, M2FTrans \cite{shi2023mftrans} and PASSION \cite{shi2024passion} — to systematically validate the impact of different modality combinations on segmentation accuracy using the same missing rate protocol. All experiments were conducted on a single NVIDIA GeForce RTX 3090 GPU using the AdamW optimizer \cite{loshchilov2017decoupled} with the following hyperparameters: 300 training epochs, initial learning rate of 2e-4, weight decay of 1e-4, and batch size of 1. The hyperparameters $\alpha_1, \lambda_1, \lambda_2, \lambda_3$, and $\lambda_4$ are initialized to values of 0.6, 2.0, 1.0, 0.5, and 0.5, respectively, following comprehensive grid search optimization.

\subsection{Evaluation on BraTS2020}
\textbf{Quantitative evaluation.} On the BraTS2020 dataset, we adopt mmFormer \cite{zhang2022mmformer} as the baseline model to systematically compare different modality rebalancing approaches (See Table \ref{tab:brats&myo}). The baseline demonstrates significantly better performance under Perfect Data Training (PDT) conditions compared to Imbalanced Data Training (IDT), highlighting the increased challenges of segmentation under imbalanced modality missing rates. The ModDrop \cite{xiao2020audiovisual} method even exhibits marginal performance degradation relative to the baseline, as its effectiveness relies heavily on precise control of dropout probabilities that cannot adapt to heterogeneous missing rates. While PMR \cite{fan2023pmr} achieves slight performance improvements over the baseline, its prototype-based approach suffers from limitations: the prototypes require sufficiently balanced and abundant samples to avoid estimation bias in medical imaging scenarios with inherent class imbalances (e.g., sparse lesion regions). This leads to suboptimal lesion segmentation performance. In contrast, PASSION \cite{shi2024passion} demonstrates notable performance gains through self-distillation and preference-aware regularization for adaptive modality learning alignment. However, it still emphasizes output probability mimicking while neglecting critical structured relationships essential for cross-modal knowledge transfer, such as spatial dependencies and pathological topological patterns. Crucially, existing modality rebalancing methods fail to address the heterogeneous modality contributions issue (e.g., Flair and T2 modalities contain significantly more diagnostic information than T1/T1c). Without dynamic weighting mechanisms for heterogeneous modality contributions, suboptimal multi-modal fusion features are generated, resulting in performance degradation. The proposed DMAF-Net comprehensively addresses both imbalanced modality missing rates and heterogeneous modality contributions through dynamic modality fusion, relation and prototype distillation, and dynamic training monitoring strategy. It achieves consistent performance improvements across all missing rate scenarios, demonstrating superior modality balancing capability in IDT conditions.

\begin{figure}[!t]
\centerline{\includegraphics[width=\columnwidth]{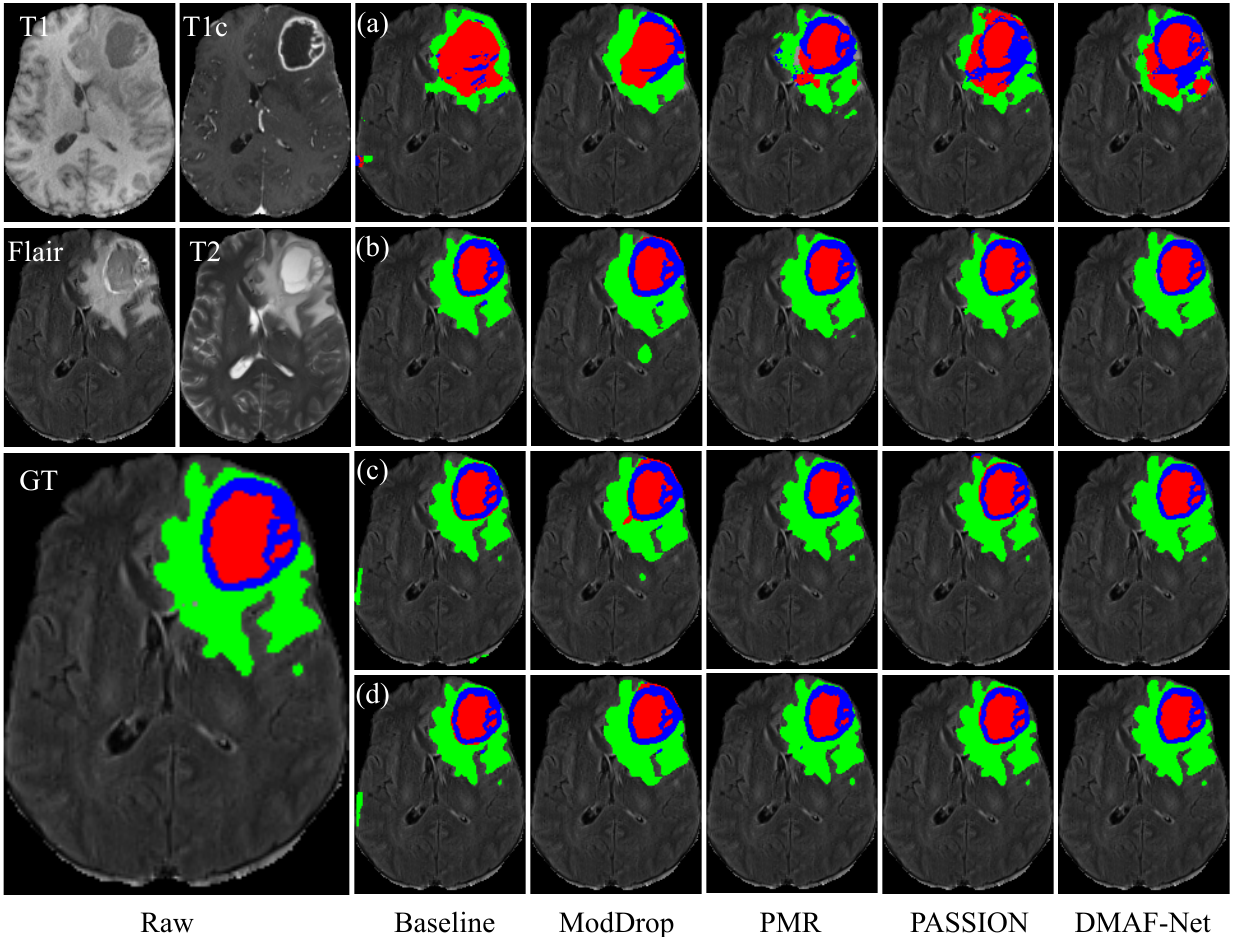}}
\caption{Segmentation visualization with MR=0.2/0.4/0.6/0.8 (T1/T1c/FLAIR/T2) on BraTS2020. Panels (a-d): T2, T1c+T2, T1c+FLAIR+T2, and all four modalities.}
\label{fig3}
\end{figure}

\textbf{Qualitative Evaluation.} Fig. \ref{fig3} presents qualitative comparisons of different modality rebalancing methods under various modality combinations on BraTS2020. When only T2 modality is available, segmentation results show significant limitations, while performance improves substantially with additional modalities, illustrating the inherent advantages of multi-modal learning over uni-modal approaches. Notably, ModDrop and PMR exhibit numerous false positives due to unaddressed imbalance in modality missing rates. In contrast, our DMAF-Net effectively suppresses false positives and produces the most accurate segmentation results.

\subsection{Evaluation on MyoPS2020}
\textbf{Quantitative evaluation.} The quantitative comparison results on MyoPS2020 are summarized in Table \ref{tab:brats&myo}. Compared to BraTS2020, MyoPS2020 exhibits less pronounced information disparity among modalities with closer semantic contributions, eliminating dominant modality effects. Consequently, the impact of imbalanced modality missing rates on segmentation performance is relatively diminished. As a result, all modality rebalancing methods demonstrate significant performance improvements over the baseline. Among these, the proposed DMAF-Net achieves the best performance, validating its superior feature extraction and fusion capabilities in medical segmentation tasks.
	
\begin{figure}[!t]
\centerline{\includegraphics[width=\columnwidth]{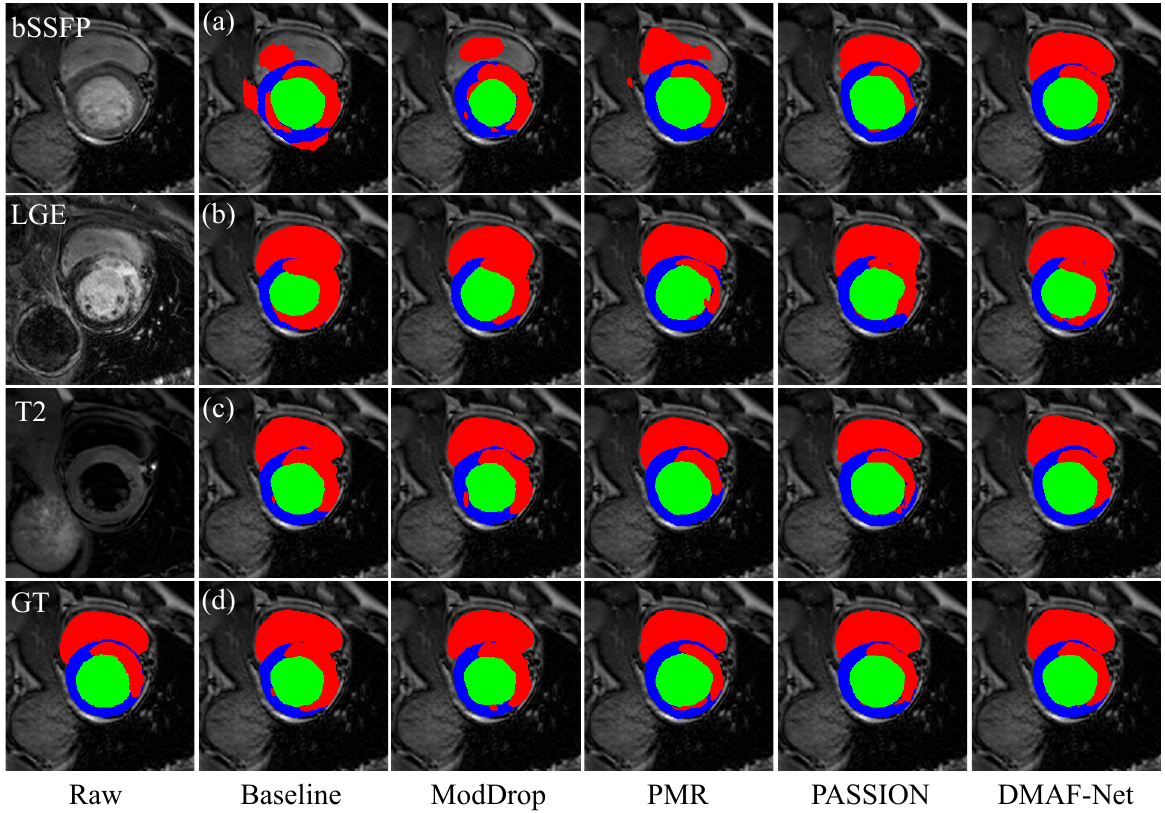}}
\caption{Segmentation visualization with MR=0.3/0.5/0.7 (bSSFP/LGE/T2) on MyoPS2020. Panels (a-d): T2, LGE, bSSFP+T2, and bSSFP+LGE+T2.}
\label{fig4}
\end{figure}

\textbf{Qualitative Evaluation.} Fig. \ref{fig4} presents qualitative comparisons of different methods under various modality combinations on MyoPS2020. Due to the balanced semantic contributions across modalities with minimal information disparity, segmentation results show smaller performance variations across different modality configurations. Compared to the baseline, all modality rebalancing methods yield more accurate segmentations, with our DMAF-Net exhibiting the fewest false positives and achieving the most optimal results. This further demonstrates the effectiveness of our dynamic modality fusion strategy in handling balanced yet challenging medical imaging scenarios.

\begin{table*}
    \caption{Quantitative evaluation of different backbone networks on BraTS2020 under varying missing modality combinations, with $MR$ set to (0.2, 0.4, 0.6, 0.8) for IDT (i.e., imperfect data training) and (0, 0, 0, 0) for PDT (i.e., perfect data training). The results in \textcolor{blue}{blue} and \textcolor{red}{red} are generated by methods that consider IDT scenarios and those that do not, respectively.}
    \label{tab:backbones}
    \resizebox{\textwidth}{!}{
        \begin{tabular}{c|c|l|cccccccccccccccc}
            \hline
            \multirow{4}{*}{Type}   & \multirow{4}{*}{Setting}    & T1    & \CIRCLE    & \Circle    & \Circle     & \Circle     & \CIRCLE    & \CIRCLE      & \CIRCLE     & \Circle     & \Circle      & \Circle       & \CIRCLE       & \CIRCLE     & \CIRCLE      & \Circle       & \CIRCLE      & \multirow{4}{*}{Avg.} \\
            & &  T1c        & \Circle               & \CIRCLE               & \Circle               & \Circle               & \CIRCLE               & \Circle               & \Circle               & \CIRCLE               & \CIRCLE               & \Circle               & \CIRCLE               & \CIRCLE               &  \Circle               & \CIRCLE               & \CIRCLE               &                      \\
            & & Flair       & \Circle               & \Circle               & \CIRCLE               & \Circle               & \Circle               & \CIRCLE               & \Circle               & \CIRCLE               & \Circle               & \CIRCLE               & \CIRCLE               & \Circle               & \CIRCLE               & \CIRCLE               & \CIRCLE               &                       \\
            & & T2        & \Circle               & \Circle               & \Circle               & \CIRCLE               & \Circle               & \Circle               & \CIRCLE               & \Circle               & \CIRCLE               & \CIRCLE               & \Circle               & \CIRCLE               & \CIRCLE               & \CIRCLE               & \CIRCLE               &                       \\
            \hline
            \multirow{10}{*}{WT} & \multirow{5}{*}{$PDT$}     
                 & RFNet\cite{ding2021rfnet}   &69.42 &70.38 &83.24 &83.29 &75.17 &86.41 &85.93 &86.15 &82.83 &85.53 &87.66 &86.40 &88.58 &88.34 &89.32 &83.24  \\
            &    & mmFormer\cite{zhang2022mmformer} &70.71 &70.76 &84.35 &84.32 &75.84 &87.50 &85.86 &87.34 &86.39 &87.81 &88.94 &87.45 &88.79 &89.35 &89.77 &84.35     \\
            &    & M2FTrans\cite{shi2023mftrans} &68.87 &68.34 &80.73 &81.15 &74.35 &86.49 &84.13 &84.84 &83.73 &87.77 &87.51 &85.12 &88.59 &88.26 &88.96 &82.59    \\
            \cline{3-19}
            &    & PASSION\cite{shi2024passion}  &70.51 &70.69 &84.22 &83.58 &76.30 &87.71 &85.62 &86.90 &86.26 &87.83 &88.65 &88.00 &88.82 &89.36 &89.74 &84.28  \\
            &    & DMAF-Net     &72.92 &70.64 &84.94 &83.64 &77.57 &87.60 &85.86 &86.90 &86.67 &88.25 &88.30 &88.27 &89.53 &89.91 &89.60 &84.71  \\
            \cline{2-19}
            & \multirow{5}{*}{$IDT$}    
                 & RFNet\cite{ding2021rfnet}    & 70.37 &70.54 &78.39 &74.62 &76.63 &85.63 &80.38 &83.91 &78.48 &83.97 &87.04 &82.54 &86.80 &85.96 &87.44 &\textcolor{blue}{80.85}     \\
            &    & mmFormer\cite{zhang2022mmformer} &66.64 &68.76 &78.81 &72.85 &77.11 &83.60 &82.42 &84.54 &78.87 &83.94 &87.20 &84.97 &86.67 &85.94 &88.06 &\textcolor{blue}{80.69}    \\
            &    & M2FTrans\cite{shi2023mftrans} &52.73 &69.94 &80.95 &72.10 &76.96 &86.61 &81.85 &84.97 &79.60 &82.37 &87.42 &83.87 &86.87 &85.62 &87.28 &\textcolor{blue}{79.94}    \\
            \cline{3-19}
            &    & PASSION\cite{shi2024passion} &72.78 &72.73 &83.37 &80.27 &77.93 &85.93 &84.05 &87.14 &84.78 &86.85 &88.07 &86.88 &87.58 &88.44 &88.88 &\textcolor{red}{83.71} \\
            &    & DMAF-Net    &75.92 &71.64 &84.94 &79.27 &78.82 &88.59 &85.55 &88.04 &84.96 &87.25 &89.30 &86.57 &88.91 &88.41 &89.20 &\textcolor{red}{84.49}    \\
            
            \cline{2-19}
            \hline
            \multirow{10}{*}{TC} & \multirow{5}{*}{$PDT$}  
                 & RFNet\cite{ding2021rfnet}    &57.64 &75.25 &62.19 &64.76 &79.56 &68.17 &68.28 &78.27 &79.88 &68.72 &79.78 &81.28 &70.72 &79.86 &80.91 &73.02    \\
            &    & mmFormer\cite{zhang2022mmformer} &54.17 &74.33 &59.68 &61.90 &78.44 &65.23 &64.55 &77.83 &79.52 &66.60 &78.87 &80.22 &67.83 &79.41 &79.90 &71.22     \\
            &    & M2FTrans\cite{shi2023mftrans} &55.80 &72.57 &56.71 &62.36 &77.48 &65.58 &66.17 &75.86 &78.57 &66.62 &78.45 &80.06 &68.21 &78.05 &80.20 &71.63    \\
            \cline{3-19}
            &    & PASSION\cite{shi2024passion}  &55.47 &73.11 &59.50 &61.76 &78.21 &66.17 &65.98 &76.33 &78.74 &67.06 &79.51 &80.54 &68.62 &79.60 & 79.93 &71.37    \\
            &    & DMAF-Net     &56.02 &74.31 &60.55 &63.18 &78.63 &66.31 &66.42 &76.85 &79.98 &67.34 &79.44 &81.06 &68.84 &80.30 &80.47 &71.98   \\
            \cline{2-19}
            & \multirow{5}{*}{$IDT$}   
                 & RFNet\cite{ding2021rfnet}    &54.87 &72.24 &49.75 &45.28 &79.04 &62.24 &57.58 &75.32 &76.41 &55.54 &77.39 &79.17 &63.52 &75.75 &78.76 &\textcolor{blue}{66.86}   \\
            &    & mmFormer\cite{zhang2022mmformer} &52.36 &71.18 &51.82 &48.35 &80.02 &59.39 &57.07 &75.56 &77.34 &56.49 &77.56 &79.70 &62.62 &76.13 &78.42 &\textcolor{blue}{66.93}  \\
            &    & M2FTrans\cite{shi2023mftrans} &34.83 &72.32 &51.27 &38.72 &79.86 &61.43 &56.85 &75.72 &77.23 &51.21 &79.20 &80.53 &62.86 &75.03 &79.81 &\textcolor{blue}{65.12}  \\
            \cline{3-19}
            &    & PASSION\cite{shi2024passion} &52.84 &73.45 &58.15 &51.56 &77.80 &63.12 &62.77 &76.16 &74.97 &58.16 &77.65 &77.07 &62.88 &75.89 &77.56 &\textcolor{red}{68.00} \\
            &    & DMAF-Net    &59.48 &75.18 &52.86 &51.31 &78.40 &66.31 &60.97 &78.86 &77.93 &59.76 &79.64 &78.95 &65.95 & 9.17 &79.55 &\textcolor{red}{69.62}   \\
            \cline{2-19}
            \hline
            \multirow{10}{*}{ET} & \multirow{5}{*}{$PDT$}
                 & RFNet\cite{ding2021rfnet}    &30.86 &64.69 &35.56 &39.17 &67.45 &40.12 &42.40 &67.27 &68.07 &42.51 &68.36 &70.12 &44.06 &67.65 &68.53 &54.45   \\
            &    & mmFormer\cite{zhang2022mmformer} &27.53 &66.14 &32.64 &37.09 &69.58 &37.40 &40.12 &67.09 &68.65 &41.06 &69.75 &69.71 &42.35 &68.70 &69.51 &53.82   \\
            &    & M2FTrans\cite{shi2023mftrans} &28.67 &65.19 &32.34 &38.17 &69.82 &37.42 &40.27 &67.05 &68.85 &41.27 &68.84 &69.76 &43.47 &68.31 &69.81 &53.95     \\
            \cline{3-19}
            &    & PASSION\cite{shi2024passion}  &28.47 &66.71 &31.25 &37.12 &69.58 &38.09 &40.66 &68.27 &67.86 &40.85 &69.27 &69.90 &42.94 &68.97 &69.77 &53.92   \\
            &    & DMAF-Net     &30.37 &66.97 &31.64 &36.86 &70.37 &38.20 &41.66 &68.62 &68.02 &40.33 &69.50 &70.11 &43.35 &69.06 &69.45 & 54.30   \\
            \cline{2-19}
            & \multirow{5}{*}{$IDT$}    
                 & RFNet\cite{ding2021rfnet}    &27.08 &62.74 &17.19 &24.54 &68.45 &35.87 &32.24 &66.31 &70.32 &21.17 &69.08 &68.21 &28.75 &67.81 &68.82 &\textcolor{blue}{48.51}    \\
            &    & mmFormer\cite{zhang2022mmformer} &26.98 &64.15 &19.13 &22.81 &70.39 &35.71 &32.75 &65.82 &68.63 &25.66 &68.14 &71.11 &30.42 &66.86 &69.28 &\textcolor{blue}{49.19}  \\
            &    & M2FTrans\cite{shi2023mftrans} &25.52 &64.32 &12.26 &20.12 &70.77 &33.34 &32.09 &65.17 &66.42 &23.85 &69.26 &70.82 &36.51 &66.80 &69.52 &\textcolor{blue}{48.38}   \\
            \cline{3-19}
            &    & PASSION\cite{shi2024passion}  &22.91 &68.31 &31.67 &26.85 &70.93 &36.43 &34.85 &69.81 &68.77 &34.15 &70.59 &70.13 &35.37 &68.92 &69.30 &\textcolor{red}{51.93} \\
            &    & DMAF-Net     &34.65 &67.87 &29.81 &21.41 &71.12 &41.04 &36.15 &69.86 &69.23 &29.56 &70.97 &70.71 &38.42 &68.97 &69.56 &\textcolor{red}{52.62}    \\
            \cline{2-19}
            \hline
        \end{tabular}
    }
\end{table*}

\subsection{Evaluation of Backbone Networks under Different Missing Modality Combinations}
To comprehensively analyze the impact of different modality-missing combinations on model performance, we conducted experiments using the BraTS2020 dataset. We compared DMAF-Net with state-of-the-art (SOTA) methods for incomplete multi-modal medical image segmentation under identical modality missing rates (0.2, 0.4, 0.6, 0.8). The detailed results are presented in Table \ref{tab:backbones}. In experiments, we observed that methods such as RFNet \cite{ding2021rfnet}, mmFormer \cite{zhang2022mmformer}, and M2Ftrans \cite{shi2023mftrans} — which were designed without considering modality imbalance and missing rate heterogeneity — exhibited significant performance degradation under IDT (Imperfect Data Training) conditions, despite their strong performance in PDT (Perfect Data Training) scenarios. This demonstrates their inadequacy for addressing the prevalent imbalanced modality missing rates encountered in clinical practice. In contrast, PASSION \cite{shi2024passion} and our proposed DMAF-Net, which explicitly account for modality imbalance, showed notably smaller performance drops compared to the aforementioned methods under IDT conditions. Remarkably, in WT (Whole Tumor) segmentation, DMAF-Net achieved performance metrics nearly comparable to PDT settings, underscoring the robustness of our framework in handling missing rate heterogeneity. Furthermore, our proposed DMAF-Net outperformed PASSION by 1-2\% in IDT scenarios, as evidenced by quantitative metrics. This improvement highlights the superiority of DMAF-Net’s dynamic fusion and training monitoring mechanisms in mitigating the challenges posed by modality imbalance, particularly in clinically relevant imperfect data conditions.

\subsection{Component-wise Ablation Study}
We conducted comprehensive component-wise ablation studies on the BraTS2020 and MyoPS2020 datasets, systematically evaluating combinations of the DMAF, Distination (encompassing both relation and prototype distillation), and DTM components. Results are summarized in Table \ref{tab:ablation}. Overall, introducing individual components separately yields performance improvements. However, the DMAF module alone demonstrates relatively limited gains, as it specifically addresses heterogeneous modality contributions while being ineffective for imbalanced modality missing rates. When all components are jointly integrated, the model simultaneously mitigates both modality missing rate and contribution imbalances, achieving optimal segmentation performance that significantly surpasses the baseline.
\renewcommand{\arraystretch}{1.18}
	\begin{table*}[htbp]
		\caption{Component-wise ablation study on BraTS2020 and MyoPS2020.}
		\label{tab:ablation}
		\resizebox{\textwidth}{!}{
			\begin{tabular}{ccc|cccc|cccc|cccc|cccc}
				\hline
				\multicolumn{3}{c|}{\multirow{2}{*}{Component}} & \multicolumn{8}{c|}{BraTS2020 $MR=(0.2, 0.4, 0.6, 0.8)$}  & \multicolumn{8}{c}{MyoPS2020 $MR=(0.3,0.5,0.7)$}   \\
				\cline{4-19}
				\multicolumn{3}{c|}{}    & \multicolumn{4}{c|}{DSC {[}\%{]} $\uparrow$}      & \multicolumn{4}{c|}{HD {[}mm{]} $\downarrow$}       & \multicolumn{4}{c|}{DSC {[}\%{]} $\uparrow$}      & \multicolumn{4}{c}{HD {[}mm{]} $\downarrow$}                                \\
				\hline
				$\mathrm{DMAF}$      & $\mathrm{DISTINATION}$      & $\mathrm{DTM}$    & WT             & TC             & ET             & Avg.            & WT             & TC             & ET            & Avg.            & LVB            & RVB            & MYO            & Avg.            & LVB            & RVB            & MYO            & Avg.            \\
				\hline
				\Circle    & \Circle    & \Circle     & 80.16    & 66.26    & 50.31    & 65.58    & 26.14    & 21.47    & 15.17   & 20.93    & 77.43    & 56.74    & 71.22    & 68.46    & 19.78    & 23.89    & 21.72    & 21.80   \\
				\CIRCLE    & \Circle    & \Circle     & 82.17    & 67.93    & 51.74    & 67.28    & 17.87    & 16.83    & 14.70   & 16.47    & 78.71    & 58.60    & 74.44    & 70.58    & 17.74    & 28.73    & 18.83    & 21.77    \\
				\Circle    & \CIRCLE    & \Circle     & 82.68    & 68.22    & 51.95    & 67.61    & 17.02    & 15.60    & 13.18   & 15.27    & 79.96    & 59.45    & 75.52    & 71.64    & 15.16    & 26.38    & 17.66    & 19.73   \\
				\Circle    & \Circle    & \CIRCLE     & 82.57    & 68.06    & 51.82    & 67.48    & 17.53    & 16.11    & 13.53   & 15.72    & 79.23    & 58.53    & 75.68    & 71.15    & 16.33    & 26.94    & 18.08    & 20.45   \\
				\CIRCLE    & \CIRCLE    & \Circle     & 84.20    & 69.31    & 52.29    & 68.60    & 12.89    & 13.15    & 12.20   & 12.75    & 80.68    & 62.40    & 75.86    & 72.98    & 13.43    & 24.40    & 14.71    & 17.51   \\
				\CIRCLE    & \Circle    & \CIRCLE     & 84.28    & 69.39    & 52.35    & 68.67    & 12.91    & 13.09    & 12.12   & 12.71    & 80.59    & 62.09    & 75.43    & 72.80    & 13.98    & 25.29    & 15.26    & 18.18   \\
                \Circle    & \CIRCLE    & \CIRCLE     & 84.26    & 69.43    & 52.39    & 68.69    & 12.82    & 13.01    & 12.13   & 12.65    & 81.06    & 63.56    & 76.05    & 73.56    & 12.69    & 23.62    & 14.47    & 16.96   \\
                \CIRCLE    & \CIRCLE    & \CIRCLE     & 84.49    & 69.62    & 52.62    & 68.91    & 11.56    & 12.77    & 11.86   & 12.06    & 81.38    & 64.07    & 76.74    & 74.05    & 11.01    & 22.88    & 12.72    & 15.54   \\
				\hline
			\end{tabular}
		}
	\end{table*}

\section{Conclusion}\label{sec5}
This paper proposes a Dynamic Modality Balancing Framework (DMAF-Net) for incomplete multi-modal medical image segmentation, aiming to address the challenges of multi-modal feature fusion and parameter optimization under imbalanced modality missing rates (IDT). To overcome the limitations of existing methods in handling modality missing rate heterogeneity and heterogeneous modality contributions, we innovatively introduce three core components: Dynamic Modality-Aware Fusion (DMAF), Relation and Prototype Distillation, and Dynamic Training Monitoring (DTM) strategy. These modules synergistically address modality imbalances by dynamically adjusting fusion weights, enforcing cross-modal semantic alignment, and adaptively stabilizing training dynamics. Comprehensive evaluations on the BraTS2020 and MyoPS2020 datasets demonstrate that our proposed DMAF-Net outperforms state-of-the-art modality rebalancing methods, confirming its superiority.

While DMAF-Net advances multi-modal segmentation, two key limitations persist: 1) The computational complexity of transformer-based cross-modal attention hinders real-time 3D medical imaging applications, and 2) Prototype Distillation remains vulnerable to annotation inaccuracies, particularly in fine-grained pathological subregions. Future work will focus on developing lightweight fusion architectures through hierarchical attention or neural architecture search to reduce computational overhead, while integrating contrastive learning with uncertainty-aware prototype alignment to enhance robustness against label noise. These optimizations aim to strengthen clinical applicability in scenarios with heterogeneous imaging protocols and imperfect annotations.

\section*{CRediT authorship contribution statement}
\textbf{Libin Lan}: Writing – review and editing, Writing – original draft, Conceptualization, Formal analysis, Investigation, Methodology, Project administration, Funding acquisition, Data curation.
\textbf{Hongxing Li}: Writing – original draft, Writing – review and editing, Methodology, Visualization, Software, Investigation.
\textbf{Zunhui Xia}: Validation, Writing – review and editing. 
\textbf{Yudong Zhang}: Writing – review and editing, Formal analysis, Conceptualization.

\section*{Declaration of competing interest}
The authors declare that they have no known competing financial interests or personal relationships that could have appeared to influence the work reported in this paper.

\section*{Acknowledgments} 
This work described in this paper was supported in part by the Scientific Research Foundation of Chongqing University of Technology under Grant 2021ZDZ030 and in part by the Youth Project of Science and Technology Research Program of Chongqing Education Commission of China under Grant KJQN202301145.

\vspace{-1em}
\bibliographystyle{elsarticle-num}
\bibliography{ref}

\begin{thebibliography}{10}
\expandafter\ifx\csname url\endcsname\relax
  \def\url#1{\texttt{#1}}\fi
\expandafter\ifx\csname urlprefix\endcsname\relax\def\urlprefix{URL }\fi
\expandafter\ifx\csname href\endcsname\relax
  \def\href#1#2{#2} \def\path#1{#1}\fi

\bibitem{bakas2017advancing}
S.~Bakas, H.~Akbari, A.~Sotiras, M.~Bilello, M.~Rozycki, J.~S. Kirby, J.~B. Freymann, K.~Farahani, C.~Davatzikos, Advancing the cancer genome atlas glioma mri collections with expert segmentation labels and radiomic features, Sci. Data. 4~(1) (2017) 1--13.

\bibitem{biondetti2021pet}
P.~Biondetti, M.~G. Vangel, R.~M. Lahoud, F.~S. Furtado, B.~R. Rosen, D.~Groshar, L.~G. Canamaque, L.~Umutlu, E.~W. Zhang, U.~Mahmood, et~al., Pet/mri assessment of lung nodules in primary abdominal malignancies: sensitivity and outcome analysis, Eur. J. Nucl. Med. Mol. I. 48 (2021) 1976--1986.

\bibitem{chen2025magnetic}
J.~Chen, J.~Cui, Y.~Yu, X.~Wang, Magnetic resonance imaging volume segmentation for breast cancer surgery, Biomed. Signal Process. Control. 110 (2025) 108131.

\bibitem{ding2021rfnet}
Y.~Ding, X.~Yu, Y.~Yang, Rfnet: Region-aware fusion network for incomplete multi-modal brain tumor segmentation, in: Proc. IEEE Int. Conf. Comput. Vision., 2021, pp. 3975--3984.

\bibitem{ma2025dmfusion}
G.~Ma, X.~Qiu, X.~Tan, Dmfusion: A dual-branch multi-scale feature fusion network for medical multi-modal image fusion, Biomed. Signal Process. Control. 105 (2025) 107572.

\bibitem{krupa2015artifacts}
K.~Krupa, M.~Bekiesi{\'n}ska-Figatowska, Artifacts in magnetic resonance imaging, POL. J. RADIOL 80 (2015) 93.

\bibitem{graves2013body}
M.~J. Graves, D.~G. Mitchell, Body mri artifacts in clinical practice: a physicist's and radiologist's perspective, J. MAGN. RESON. IMAGING. 38~(2) (2013) 269--287.

\bibitem{havaei2016hemis}
M.~Havaei, N.~Guizard, N.~Chapados, Y.~Bengio, Hemis: Hetero-modal image segmentation, in: Proc. Int. Conf. Med. Image Comput. Comput.-Assisted Intervention., Springer, 2016, pp. 469--477.

\bibitem{wang2023a2fseg}
Z.~Wang, Y.~Hong, A2fseg: Adaptive multi-modal fusion network for medical image segmentation, in: Proc. Int. Conf. Med. Image Comput. Comput.-Assisted Intervention., Springer, 2023, pp. 673--681.

\bibitem{shi2024passion}
J.~Shi, C.~Shang, Z.~Sun, L.~Yu, X.~Yang, Z.~Yan, Passion: Towards effective incomplete multi-modal medical image segmentation with imbalanced missing rates, in: Proc. ACM Int. Conf. Multimed., 2024, pp. 456--465.

\bibitem{goodfellow2020generative}
I.~Goodfellow, J.~Pouget-Abadie, M.~Mirza, B.~Xu, D.~Warde-Farley, S.~Ozair, A.~Courville, Y.~Bengio, Generative adversarial networks, COMMUN. ACM. 63~(11) (2020) 139--144.

\bibitem{meng2024multi}
X.~Meng, K.~Sun, J.~Xu, X.~He, D.~Shen, Multi-modal modality-masked diffusion network for brain mri synthesis with random modality missing, {IEEE} Trans. Med. Imag. (2024).

\bibitem{muller2023multimodal}
G.~M{\"u}ller-Franzes, J.~M. Niehues, F.~Khader, S.~T. Arasteh, C.~Haarburger, C.~Kuhl, T.~Wang, T.~Han, T.~Nolte, S.~Nebelung, et~al., A multimodal comparison of latent denoising diffusion probabilistic models and generative adversarial networks for medical image synthesis, Sci. Rep. 13~(1) (2023) 12098.

\bibitem{zhang2024unified}
Y.~Zhang, C.~Peng, Q.~Wang, D.~Song, K.~Li, S.~K. Zhou, Unified multi-modal image synthesis for missing modality imputation, {IEEE} Trans. Med. Imag. (2024).

\bibitem{zhang2024brain}
Y.~Zhang, G.~Liu, R.~Bao, L.~Zhan, P.~Thompson, H.~Huang, Brain image synthesis using incomplete multimodal data, in: Proc. IEEE Int. Symp. Biomed. Imaging, IEEE, 2024, pp. 1--5.

\bibitem{wang2023prototype}
S.~Wang, Z.~Yan, D.~Zhang, H.~Wei, Z.~Li, R.~Li, Prototype knowledge distillation for medical segmentation with missing modality, in: Proc. IEEE Int. Conf. Acoust. Speech Signal Process., IEEE, 2023, pp. 1--5.

\bibitem{wang2023learnable}
H.~Wang, C.~Ma, J.~Zhang, Y.~Zhang, J.~Avery, L.~Hull, G.~Carneiro, Learnable cross-modal knowledge distillation for multi-modal learning with missing modality, in: Proc. Int. Conf. Med. Image Comput. Comput.-Assisted Intervention., Springer, 2023, pp. 216--226.

\bibitem{zhang2022mmformer}
Y.~Zhang, N.~He, J.~Yang, Y.~Li, D.~Wei, Y.~Huang, Y.~Zhang, Z.~He, Y.~Zheng, mmformer: Multimodal medical transformer for incomplete multimodal learning of brain tumor segmentation, in: Proc. Int. Conf. Med. Image Comput. Comput.-Assisted Intervention., Springer, 2022, pp. 107--117.

\bibitem{li2024deformation}
Z.~Li, Y.~Zhang, H.~Li, Y.~Chai, Y.~Yang, Deformation-aware and reconstruction-driven multimodal representation learning for brain tumor segmentation with missing modalities, BSPC 91 (2024) 106012.

\bibitem{wang2023multi}
H.~Wang, Y.~Chen, C.~Ma, J.~Avery, L.~Hull, G.~Carneiro, Multi-modal learning with missing modality via shared-specific feature modelling, in: Proc. IEEE Int. Conf. Comput. Vision Pattern Recognit., 2023, pp. 15878--15887.

\bibitem{wei2024fly}
Y.~Wei, D.~Hu, H.~Du, J.-R. Wen, On-the-fly modulation for balanced multimodal learning, {IEEE} Trans. Pattern Anal. Mach. Intell. (2024).

\bibitem{peng2022balanced}
X.~Peng, Y.~Wei, A.~Deng, D.~Wang, D.~Hu, Balanced multimodal learning via on-the-fly gradient modulation, in: Proc. IEEE Int. Conf. Comput. Vision Pattern Recognit., 2022, pp. 8238--8247.

\bibitem{fan2023pmr}
Y.~Fan, W.~Xu, H.~Wang, J.~Wang, S.~Guo, Pmr: Prototypical modal rebalance for multimodal learning, in: Proc. IEEE Int. Conf. Comput. Vision Pattern Recognit., 2023, pp. 20029--20038.

\bibitem{yang2024facilitating}
Y.~Yang, F.~Wan, Q.-Y. Jiang, Y.~Xu, Facilitating multimodal classification via dynamically learning modality gap, Proc. Adv. Neural Inf. Proces. Syst. 37 (2024) 62108--62122.

\bibitem{zhang2024multimodal}
X.~Zhang, J.~Yoon, M.~Bansal, H.~Yao, Multimodal representation learning by alternating unimodal adaptation, in: Proc. IEEE Int. Conf. Comput. Vision Pattern Recognit., 2024, pp. 27456--27466.

\bibitem{wei2024enhancing}
Y.~Wei, R.~Feng, Z.~Wang, D.~Hu, Enhancing multimodal cooperation via sample-level modality valuation, in: Proc. IEEE Int. Conf. Comput. Vision Pattern Recognit., 2024, pp. 27338--27347.

\bibitem{zhang2024unleashing}
Y.~Zhang, Z.~Chen, L.~Liang, H.~Chen, W.~Zhang, Unleashing the power of imbalanced modality information for multi-modal knowledge graph completion, in: Proc. Jt. Int. Conf. Comput. Linguist., 2024, pp. 17120--17130.

\bibitem{sun2024redcore}
J.~Sun, X.~Zhang, S.~Han, Y.-P. Ruan, T.~Li, Redcore: Relative advantage aware cross-modal representation learning for missing modalities with imbalanced missing rates, in: Proc. AAAI Conf. Artif. Intell., Vol.~38, 2024, pp. 15173--15182.

\bibitem{ho2020denoising}
J.~Ho, A.~Jain, P.~Abbeel, Denoising diffusion probabilistic models, Proc. Adv. Neural Inf. Proces. Syst. 33 (2020) 6840--6851.

\bibitem{ma2021smil}
M.~Ma, J.~Ren, L.~Zhao, S.~Tulyakov, C.~Wu, X.~Peng, Smil: Multimodal learning with severely missing modality, in: Proc. AAAI Conf. Artif. Intell., Vol.~35, 2021, pp. 2302--2310.

\bibitem{zhang2023motion}
L.~Zhang, B.~Jiang, Q.~Chen, L.~Wang, K.~Zhao, Y.~Zhang, R.~Vliegenthart, X.~Xie, Motion artifact removal in coronary ct angiography based on generative adversarial networks, EUR. RADIOL. 33~(1) (2023) 43--53.

\bibitem{ahmad2022new}
W.~Ahmad, H.~Ali, Z.~Shah, S.~Azmat, A new generative adversarial network for medical images super resolution, Sci. Rep. 12~(1) (2022) 9533.

\bibitem{wu2024prototype}
A.~Wu, J.~Yu, Y.~Wang, C.~Deng, Prototype-decomposed knowledge distillation for learning generalized federated representation, {IEEE} Trans. Multimedia (2024).

\bibitem{wei2023mmanet}
S.~Wei, C.~Luo, Y.~Luo, Mmanet: Margin-aware distillation and modality-aware regularization for incomplete multimodal learning, in: Proc. IEEE Int. Conf. Comput. Vision Pattern Recognit., 2023, pp. 20039--20049.

\bibitem{xiao2020audiovisual}
F.~Xiao, Y.~J. Lee, K.~Grauman, J.~Malik, C.~Feichtenhofer, Audiovisual slowfast networks for video recognition, arXiv preprint arXiv:2001.08740 (2020).

\bibitem{dou2020unpaired}
Q.~Dou, Q.~Liu, P.~A. Heng, B.~Glocker, Unpaired multi-modal segmentation via knowledge distillation, {IEEE} Trans. Med. Imag. 39~(7) (2020) 2415--2425.

\bibitem{salguero2024data}
J.~Salguero, P.~Prasanna, G.~Corredor, A.~Cruz-Roa, D.~Becerra, E.~Romero, Data distillation in computational pathology by choosing few representants of the original variance: A use case in ovarian cancer, EXPERT. SYST. APPL. 245 (2024) 123028.

\bibitem{wang2025distilling}
K.~Wang, F.~Zheng, D.~Guan, J.~Liu, J.~Qin, Distilling heterogeneous knowledge with aligned biological entities for histological image classification, Pattern Recognit. 160 (2025) 111173.

\bibitem{shi2023mftrans}
J.~Shi, L.~Yu, Q.~Cheng, X.~Yang, K.-T. Cheng, Z.~Yan, Mftrans: Modality-masked fusion transformer for incomplete multi-modality brain tumor segmentation, {IEEE} J. Biomed. Health Inform. 28~(1) (2023) 379--390.

\bibitem{qian2025dyncim}
C.~Qian, K.~Han, J.~Wang, Z.~Yuan, R.~Qian, C.~Lyu, J.~Chen, Z.~Liu, Dyncim: Dynamic curriculum for imbalanced multimodal learning, arXiv preprint arXiv:2503.06456 (2025).

\bibitem{li2023boosting}
H.~Li, X.~Li, P.~Hu, Y.~Lei, C.~Li, Y.~Zhou, Boosting multi-modal model performance with adaptive gradient modulation, in: Proc. IEEE Int. Conf. Comput. Vision., 2023, pp. 22214--22224.

\bibitem{wei2024diagnosing}
Y.~Wei, S.~Li, R.~Feng, D.~Hu, Diagnosing and re-learning for balanced multimodal learning, in: Proc. Eur. Conf. Comput. Vis., Springer, 2024, pp. 71--86.

\bibitem{morales2024exponential}
D.~Morales-Brotons, T.~Vogels, H.~Hendrikx, Exponential moving average of weights in deep learning: Dynamics and benefits, arXiv preprint arXiv:2411.18704 (2024).

\bibitem{menze2014multimodal}
B.~H. Menze, A.~Jakab, S.~Bauer, J.~Kalpathy-Cramer, K.~Farahani, J.~Kirby, Y.~Burren, N.~Porz, J.~Slotboom, R.~Wiest, et~al., The multimodal brain tumor image segmentation benchmark (brats), {IEEE} Trans. Med. Imag. 34~(10) (2014) 1993--2024.

\bibitem{qiu2023myops}
J.~Qiu, L.~Li, S.~Wang, K.~Zhang, Y.~Chen, S.~Yang, X.~Zhuang, Myops-net: Myocardial pathology segmentation with flexible combination of multi-sequence cmr images, MED. IMAGE. ANAL. 84 (2023) 102694.

\bibitem{zhuang2018multivariate}
X.~Zhuang, Multivariate mixture model for myocardial segmentation combining multi-source images, {IEEE} Trans. Pattern Anal. Mach. Intell. 41~(12) (2018) 2933--2946.

\bibitem{loshchilov2017decoupled}
I.~Loshchilov, F.~Hutter, Decoupled weight decay regularization, arXiv preprint arXiv:1711.05101 (2017).

\end{thebibliography}

\end{document}